\documentclass[lettersize,journal]{IEEEtran}
\usepackage[dvipsnames]{xcolor}
\usepackage{graphicx}
\usepackage{subcaption}
\usepackage{tabularx}
\usepackage{amsfonts} 
\usepackage{graphicx}
\usepackage{cite}
\usepackage{amsmath}
\usepackage{comment}
\usepackage{enumerate}

\usepackage[dvipsnames]{xcolor}
\newcommand{\vj}[1]{\textcolor{black}{#1}}

\usepackage{color}
\definecolor{myGreen}{RGB}{255, 0, 0}

\begin{document}
\title{\vj{Model-free Learning of Corridor Clearance: \\A Near-term Deployment Perspective}}

\author{Dajiang Suo$^{1*}$, Vindula Jayawardana$^{2*}$, Cathy Wu$^{3}$
        
\thanks{$^{*}$Suo and Jayawardana contribute equally to this work.}
\thanks{This research was partially supported by the MIT-IBM Watson AI Lab, the US DOT's Federal Highway Administration and Utah Department of Transportation under project number F-ST99(783), the MIT Amazon Science Hub, and the National Science Foundation (NSF) under grant number 2149548.}
\thanks{$^{1}$D. Suo is with the Polytechnic School in the Ira A. Fulton Schools of Engineering, Arizona State University, Tempe, AZ 85281, USA
        {\tt\small dajiang.suo@asu.edu}}%
\thanks{$^{2}$V. Jayawardana is with the Department of Electrical Engineering and Computer Science and Laboratory for Information \& Decision Systems, Massachusetts Institute of Technology, Cambridge, MA 02139, USA
        {\tt\small vindula@mit.edu}}%

\thanks{$^{3}$C. Wu is with the Laboratory for Information \& Decision Systems, Department of Civil and Environmental Engineering and Institute of Data Systems and Society, Massachusetts Institute of Technology, Cambridge, MA 02139, USA
        {\tt\small cathywu@mit.edu}}%

\thanks{Corresponding author: Dajiang Suo (dajiang.suo@asu.edu)}
}

\markboth{This paper has been accepted for publication in IEEE Transaction on Intelligent Transportation Systems}%
{Shell \MakeLowercase{\textit{et al.}}: A Sample Article Using IEEEtran.cls for IEEE Journals}


\maketitle

\begin{abstract}
\vj{
An emerging public health application of connected and automated vehicle (CAV) technologies is to reduce response times of emergency medical service (EMS) by indirectly coordinating traffic. Therefore, in this work we study the CAV-assisted corridor clearance for EMS vehicles from a short term deployment perspective. Existing research on this topic often overlooks the impact of EMS vehicle disruptions on regular traffic, assumes 100\% CAV penetration, relies on real-time traffic signal timing data and queue lengths at intersections, and makes various assumptions about traffic settings when deriving optimal model-based CAV control strategies. However, these assumptions pose significant challenges for near-term deployment and limit the real-world applicability of such methods. To overcome these challenges and enhance real-world applicability in near-term, we propose a model-free approach employing deep reinforcement learning (DRL) for designing CAV control strategies, showing its reduced overhead in designing and greater scalability and performance compared to model-based methods. Our qualitative analysis highlights the complexities of designing scalable EMS corridor clearance controllers for diverse traffic settings in which DRL controller provides ease of design compared to the model-based methods. In numerical evaluations, the model-free DRL controller outperforms the model-based counterpart by improving traffic flow and even improving EMS travel times in scenarios when a single CAV is present. Across 19 considered settings, the learned DRL controller excels by 25\% in reducing the travel time in six instances, achieving an average improvement of 9\%. These findings underscore the potential and promise of model-free DRL strategies in advancing EMS response and traffic flow coordination, with a focus on practical near-term deployment.}

\end{abstract}


\section{Introduction}

\IEEEPARstart{T}{he} \vj{efficient and timely response of emergency medical service (EMS) vehicles is crucial for saving lives in situations such as road incidents and natural disasters. However, the quick arrival of EMS vehicles can often be hindered by various factors, leading to potentially high fatality rates~\cite{MA2019379}. Gonzales et al.~\cite{GONZALEZ200930} show that the average emergency medical services response time for rural motor vehicle accidents involving survivors was 8.54 minutes, whereas the average response time was 10.67 minutes for incidents resulting in fatalities. This suggests a correlation between extended EMS prehospital time and elevated mortality rates.}

\vj{In an attempt to minimize response time, emergency responders may resort to driving at high speeds through city streets or areas with heavy traffic. Unfortunately, this approach can disrupt the flow of traffic and increase the risk to both nearby vehicles and vulnerable road users. Despite the development of technologies that offer emergency warning systems and traffic signal preemption capabilities, accidents involving emergency vehicles still contribute significantly to road incidents, resulting in severe injuries and fatalities. The financial toll of these incidents is substantial, with estimates suggesting an annual cost of \$35 billion in the United States alone~\cite{evdeath,blincoe2022economic} and approximately \$250 billion globally. Therefore, it is imperative for emergency responders to strike a balance between reaching incident sites promptly and minimizing the adverse impact of their response on regular traffic flow.}


\begin{figure}[tb!]
    \centering
    \includegraphics[width=0.4\textwidth]{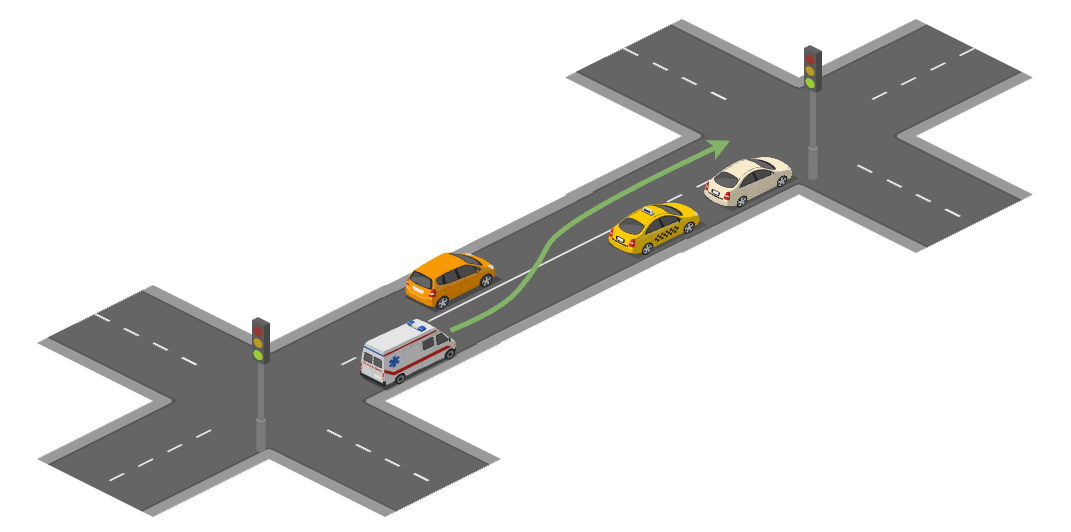}
    \caption{An illustrative scenario of EMS corridor clearance with the support of a single CAV. The CAV is shown in orange, and all other vehicles are human-driven vehicles. In assisting the EMS vehicle, the CAV blocks its following traffic, thereby creating an opportunity for the EMS vehicle to switch from a congested lane (right) to the lane that carries less traffic (left). By taking such lane-changing maneuvers, the EMS vehicle can get a head start in crossing the intersection without stopping or with fewer speed reductions, thereby reducing its travel time.}
    \label{fig:scenario}
\end{figure}


Recent studies show that connected and automated vehicle (CAV) technologies have great potential for improving the safety and efficiency of emergency responses~\cite{cavneeds}. Specifically, automated vehicles installed with wireless communication modules, such as Vehicle-to-Vehicle (V2V) communications, can be programmed to exchange information with EMS vehicles in non-line-of-sight scenarios and assist the cooperative maneuvers between EMS vehicle and non-EMS vehicles under emergency modes~\cite{hannoun2018facilitating}. Fig.~\ref{fig:scenario} shows such an example illustration of reducing EMS vehicle travel time by corridor clearance with the assistance of a single CAV near a signalized intersection. 

\vj{Despite the promising nature of CAV-enabled approaches, previous research heavily relies on strong assumptions. These assumptions can be broadly categorized into two major categories, namely 1). modeling related assumptions, and 2) control synthesis-based assumptions. Modeling-related assumptions primarily involve assuming a high CAV penetration rate and constant availability of smart road infrastructure. Control synthesis-based assumptions are primarily related to assuming a model of vehicle dynamics that is too complicated to define, such as human-driven vehicles. Moreover, due to the same model-based assumptions, most studies only consider the travel time of the EMS vehicle and do not consider the induced behavioral change to surrounding traffic. For instance, prioritizing emergency medical service vehicles at intersections frequently interrupts the traffic flow both into and out of the intersection, subsequently leading to a decrease in throughput. We realize these assumptions as too restrictive, specifically given the opportunity the CAV-assistance EMS path clearance problem promotes and by looking at the challenges to deploy them in the real world in the short term.}

\vj{Higher levels of CAV penetration are not achievable in the near future and thus are only a long-term possibility. In order to yield benefits in the short term, methodologies that explicitly take mixed traffic into consideration need to be defined. It should be noted that with a low penetration rate, the EMS path clearance problem becomes even harder as the means of system control is less compared to scenarios with higher levels of CAV penetration. Similarly, the assumptions on smart road infrastructure result in likewise limitations. These assumptions concern that the CAV platoon near an EMS vehicle always has access to comprehensive information about local traffic, including traffic flow, lane density, and speed, individual CAV statuses (e.g., speed and position), and signal phasing and timing (SPaT) information of traffic signals. The infrastructure needed for such developments is far from complete and thus limits the deployments of the previous work. }

\begin{figure*}[tb!]
  \centering
  \begin{subfigure}{0.49\textwidth}
    \centering
    \includegraphics[width=\textwidth]{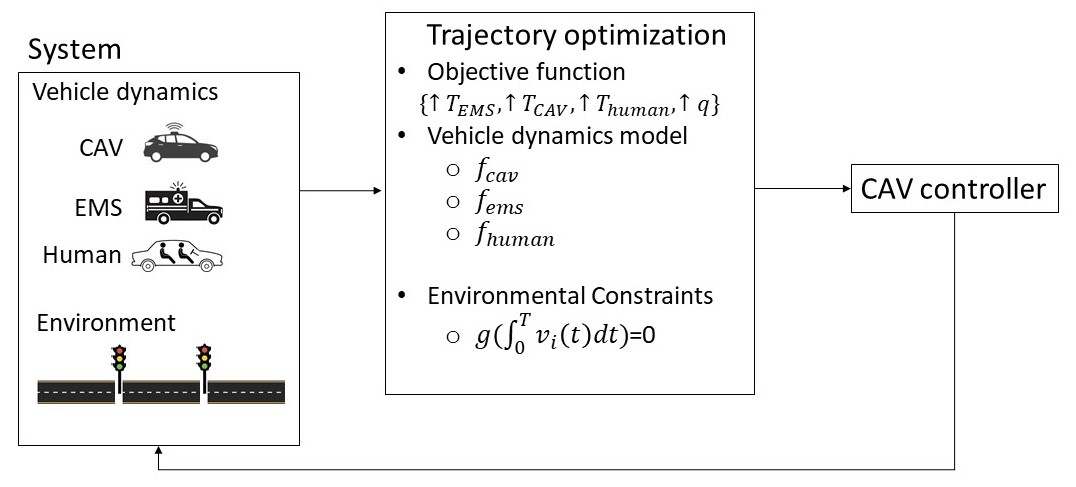}
    \caption{Trajectory optimization}
    \label{fig:model-based_compare}
  \end{subfigure}
  \begin{subfigure}{0.48\textwidth}
    \centering
    \includegraphics[width=\textwidth]{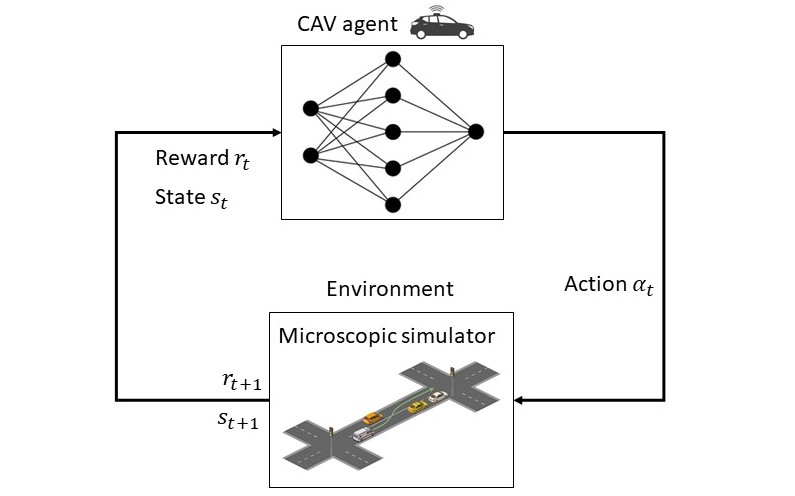}
    \caption{Deep reinforcement learning}
    \label{fig:model-free_compare}
  \end{subfigure}
  \caption{Model-based control vs. Model-free learning for EMS corridor clearance. }
  \label{fig:comparision_methods}
  \end{figure*}

\vj{The control synthesis-related assumptions, to a certain extent, overlap with the modeling-related assumptions. For example, popular assumptions such as access to SPaT messages from traffic signals in designing CAV control algorithms inherently require modeling assumptions on the availability of edge devices on the traffic signals. However, the most important control synthesis-based assumptions are based on the fact that many studies rely on a model of vehicle dynamics for control synthesis. These methods are commonly known as \textit{model-based methods}, as shown in Fig.~\ref{fig:model-based_compare}. However, it is well known that vehicle dynamics, especially human-driven vehicles, are too complicated to be specified in closed form, specifically when inter-vehicle dynamics are also considered. Thus, previous studies rely on simplified models, which may result in overestimation or underestimation of expected benefits. Furthermore, the limitations of having access to a vehicle dynamics model in the presence of signalized intersections have resulted in previous works only analyzing the benefits by considering a single intersection. Extending such analysis to multiple intersections is a tedious task that requires redefining the model and control-synthesis methods that can accommodate multiple intersections, a task that is highly unscalable. }

\vj{In this work, we aim to relax these assumptions on both fronts. First, we relax the high penetration of CAVs to a single CAV, modeling the near-term expectations. This naturally leads to modeling mixed traffic context (where both human drivers and CAV co-exist) where the system-level control is much harder due to fewer means of control enforcers (only a single CAV can enforce a control to yield system-level benefits). Second, we do not assume access to comprehensive information about local traffic, aiming for another requirement for short-term development. Instead, we only assume the information exchange between the EMS vehicle and a single CAV when they are in close vicinity. Third, most importantly, to relax the model-based assumptions, we utilize model-free deep reinforcement learning (DRL) for CAV control synthesis, as shown in Fig.~\ref{fig:model-free_compare}. We demonstrate, unlike previous model-based approaches, the scalability of our deep reinforcement learning-based method by extending our CAV controller from a single intersection to two intersections just with minor modifications. For the comparison, we also extend a previous model-based method to two intersections and show that our DRL controller outperforms the model-based method on multiple fronts despite requiring substantially fewer modifications.  }

\vj{We note that our main objective of this paper is not to devise a comprehensive CAV control synthesis strategy that generalizes to multiple CAV penetration levels nor to multiple signalized intersections. Our main focus relies on two aspects, 1) capture the requirements of the EMS path clearance problem when it comes to short-term deployment. We highlight that the stochasticity and uncertainty induced by such short-term deployment requirements make the problem more complex and different from its counterpart in long-term deployment. 2) To capture those complexities, we require more scalable and generalizable control synthesis methods and thus raise the opportunity to leverage model-free reinforcement learning as a means of control synthesis. In a broader sense, we hope our work will shed light on challenges involved in synthesizing control methods aiming for short-term deployment and raise deep reinforcement learning as a potential framework for handling such complexities. }

\vj{In light of this, our main contributions of this work involve,}
\vj{
\begin{itemize}
    \item We relax the modeling-related assumptions by only considering a single CAV in the vicinity of the EMS vehicle (the lowest penetration) and further by only considering information exchange between the CAV and EMS vehicle (as opposed to with traffic signals and nearby vehicles in the previous work)
    \item To tackle control synthesis-based challenges, we formulate the EMS corridor clearance problem at signalized intersections as a Partially Observable Markov Decision Process (POMDP), and synthesis control for CAVs when the penetration is the lowest, using model-free DRL (as opposed to defining a model of vehicle dynamics in previous work). 
    \item We benchmark the learned policies of the DRL controller against a derived model-based controller designed using shockwave theory in simulations. Our results indicate that the DRL controller outperforms the model-based controller on multiple fronts while providing greater flexibility for extending to multiple intersections.
\end{itemize}
}

\begin{table}[t]
	\caption{Notation used in the remainder of the paper}
		\begin{tabularx}{\columnwidth}{c|X}
			\hline
			\textbf{Symbol}&\textbf{Meaning} \\
			\hline
			$CAV$& connected and automated vehicle  \\
			\hline
			$EV$& EMS vehicle  \\
			\hline
			$x_L$& the optimal splitting point for the CAV platoon originally proposed in the Vanilla Jordan controller~\cite{jordan2013path}. It represents the distance from the intersection to the optimal location where the platoon in the EMS vehicle's adjacent lane should split  \\
			\hline
			$x_a$& the distance from the intersection to the actual location of the CAV when the traffic signal (S-N) turns from red to green\\
                \hline
			$x_o$& for the model-based controller designed for two-intersection corridor clearance, it represents the adjusted lane-changing point by the EMS vehicle, as opposed to the optimal splitting point proposed in vanilla Jordan controller~\cite{jordan2013path}\\
			\hline
			$x_b$& for the model-based controller designed for two-intersection corridor clearance, it represents the distance between the second intersection and the position of the CAV's leading vehicle\\
			\hline
                $l_i$& distance from the intersection to the actual location of the $i$th human-driven vehicle when the traffic signal (S-N) turns from red to green\\
			\hline
			$d$& distance from the intersection to the actual location of the EMS vehicle when the traffic signal (S-N) turns from red to green\\
			\hline
			$z$& for the model-based controller designed for two-intersection corridor clearance, it represents the distance between the stop line of the first and the second intersection\\
			\hline
			$w$& speed of queue discharging (i.e., the shockwave speed)\\
			\hline
			$V$& desired speed of the EMS vehicle (a variable subject to change and is bounded by $V_{max}$)\\
			\hline
			$v_{i}$& speed of vehicle $i$\\
			\hline
                $a_{i}$& acceleration of vehicle $i$\\
			\hline
			$v_{ev}$& speed of the EMS vehicle\\
			\hline
			$v_{cav}$& speed of the CAV\\
			\hline
			$p_{cav}$& position of the CAV\\
			\hline
			$p_{ev}$& position of the EMS vehicle\\
			\hline
		    $l_{ev}$& lane of the EMS vehicle\\
			\hline
			$l_{cav}$& lane of the CAV\\
			\hline
		    $h_{i}$& relative distance between CAV and vehicle $i$\\
			\hline
			$U$&desired speed of other vehicles (a constant) \\
			\hline
			$t_0$& time when the queue starts to discharge, i.e. signal turns green \\
			\hline
			$t_2$& time when the EMS vehicle starts to move \\
			\hline
			$t_s$& time when the EMS vehicle switch to the left lane \\
			\hline
			$t_a$& time when the CAV is first able to move, following the queue discharge \\
			\hline
			$t^{ev}_{end}$& time when the EMS vehicle has just crossed the intersection and entered the next road segment\\
			\hline
			$t^{cav}_{end}$& time when the CAV vehicle has just crossed the intersection and entered the next road segment\\
			\hline
		 \end{tabularx}
		\label{tab1}
\end{table}

\section{Background and related work}

A majority of previous work on reducing EMS response time focuses on the assignment of EMS vehicles and their routing~\cite{humagain2020systematic,fleischman2013predicting}. Additionally, there exists work that takes a microscopic view that focuses on the priority of EMS in a given road segment~\cite{hannoun2018facilitating,wu2020emergency,hannoun2021sequential}. Since the response time of an EMS vehicle is defined as the time between an emergency call is received and the EMS vehicle arrives at the emergency site~\cite{alanis2013markov}, any events that occur along the EMS vehicle trajectory can result in delays in the response time. For example, congestion in a road segment or near an intersection can increase the travel time of the EMS vehicle. Due to the importance of fine-grained interaction effects between vehicles, we consider a purely microscopic view of the problem in this article, forgoing the macroscopic view leveraged in prior works. This paper thus explores the scenario in which an EMS vehicle approaches two consecutive signalized intersections that are congested. We are primarily interested in devising strategies for CAVs to best assist the EMS vehicle in crossing the two consecutive intersections with minimum disturbances to its desired speed. 

In terms of applying CAV technologies for EMS corridor clearance near intersections, there exists a wealth of literature on infrastructure-based approaches for clearing corridors for EMS vehicles, such as traffic signal preemption by utilizing vehicle-to-infrastructure (e.g., traffic signal) communication. Wang et al. propose a traffic signal preemption strategy based on V2I technologies to discharge the flow before an EMS vehicle enters the queue~\cite{wang2013design}. Huang et al. suggest that the head green time, which accounts for the early arrivals of EMS vehicles, and the extended green time, which accounts for the late arrivals of the EMS vehicles, be added to the green light phase to avoid intersection delays~\cite{huang2011signal}. 

In addition to using smart infrastructure (e.g., traffic signals) to clear queues for the EMS vehicle near intersections, there exist alternative methods that utilize the cooperation between CAVs and the EMS vehicle. Jordan et al.~\cite{jordan2013path} propose creating a split in the vehicle queue in one lane at a critical location such that the EMS vehicle that is on the other lane can change lanes at that critical location and travel at its desired speed. While Jordan et al.'s approach is technically feasible, it is based on the strong assumption that all vehicles in the adjacent lane of the EMS vehicles are installed with V2V modules and controlled by automated controllers (i.e., 100$\%$ penetration rate of CAVs) such that they will behave exactly the way as instructed to split at an ``optimal'' location. This assumption might become invalid as long as there exist non-CAVs whose behaviors diverge from what is specified. 


In this paper, we overcome the aforementioned limitation by extending the model proposed by Jordan et al.~\cite{jordan2013path} and relaxing the assumption on CAV penetration rate. For ease of exposition, we assume that there is \textit{only one} CAV in the adjacent lane to the EMS vehicle. We take inspiration from \cite{wu2021flow}, in which a single CAV in a system of dozens of vehicles is demonstrated to eliminate stop-and-go traffic congestion and avoid intersection queueing. A key insight we leverage is that although a majority of vehicles are not directly controlled, they can be \textit{indirectly} and effectively influenced via control of the CAV, for instance, by limiting the speed of following vehicles. We additionally leverage the Flow framework introduced in~\cite{wu2021flow} to conduct our DRL-based experiments.

With recent promising developments in deep reinforcement learning, much focus has been given to learning controllers (model-free) instead of defining them explicitly (model-based). The possibility of learning a control model without a defined system dynamics model is an appealing advantage of the application of deep reinforcement learning. By leveraging the potential of DRL, Su et al.~\cite{Su2021EMVLightAD} present EMVLight, a decentralized reinforcement learning framework for simultaneous dynamic routing and traffic signal control to reduce EMS vehicle travel time. Yan et al.~\cite{Yan2021RefinedPP} propose a Deep Q learning-based path planning for EMS vehicles under different traffic control schemes. This paper proposes a DRL-based controller for the \textit{CAV} to assist EMS vehicles in passing through intersections with minimum disturbances. As such, our proposed model is different from and complementary to the existing works.

\section{Preliminaries}

\subsection{Reinforcement Learning}
\label{MDP_basic}
In designing the DRL controller, we formulate the \textit{EMS} corridor clearance problem as a finite horizon discounted Markov Decision Process (MDP) defined by the six-tuple $\mathcal{M} = \left\langle\mathcal{S},\mathcal{A}, p, r, \rho_{0}, \gamma \right\rangle$ consisting of the state space $\mathcal{S}$, the action space $\mathcal{A}$, a stochastic transition function $p : \mathcal{S} \times \mathcal{A} \rightarrow \Delta(\mathcal{S})$, a reward function $r: \mathcal{S} \times \mathcal{A} \times \mathcal{S} \rightarrow \mathbb{R}$, initial state distribution $\rho_{0} : \mathcal{S} \rightarrow \Delta(\mathcal{S})$ and a discount factor $\gamma \in [0,1]$. Here, $\Delta (\mathcal{S})$ is the probability simplex over $\mathcal{S}$. The objective then becomes searching for a policy $\pi_{\theta} : \mathcal{S} \rightarrow \Delta\mathcal{(A)}$ that will maximize the expected cumulative discounted reward over the MDP as in Equation~\ref{rl-obj}. Here $\pi_{\theta}$ is parameterized by $\theta$ and often modeled as a neural network. 

\begin{equation}
\max_{\theta} \mathbb{E}_{s_{0} \sim \rho_{0}, a_{t} \sim \pi_{\theta}\left(\cdot | s_{t}\right), s_{t+1} \sim p(\cdot | s_t,a_t)}\left[\sum_{t=0}^{T-1} \gamma^{t} r\left(s_{t}, a_{t}, s_{t+1}\right)\right]
\label{rl-obj}
\end{equation}

In this work, we use policy gradient methods to optimize for a locally optimal policy. The vanilla policy gradient algorithm~\cite{NIPS1999_464d828b} optimizes the above objective by sampling trajectories ($s_0,a_0,\cdots,s_{T-1},a_{T-1}, s_{T}$) and following the update rule, 

\begin{equation}
\theta \leftarrow \theta+\beta \sum_{t=0}^{T-1} \nabla_{\theta} \log \pi_{\theta}(s_{t}) \sum_{t^{\prime}=t}^{T-1} \gamma^{t^{\prime}-t} r\left(s_{t^{\prime}}, a_{t^{\prime}}, s_{t^{\prime}+1}\right)
\end{equation}

where $\beta$ is the learning rate. We use the Proximal Policy Optimization (PPO) algorithm~\cite{Schulman2017ProximalPO}.

\section{Modeling for EMS Corridor Clearance Problem}

\vj{In this section, we introduce the EMS corridor clearance problem in congested intersections from a microscopic perspective and discuss the modeling requirements concerning short-term deployment. }

\begin{figure*}[tb!]
\centering
\begin{subfigure}{0.46\textwidth}
  \centering
  \includegraphics[width=\textwidth]{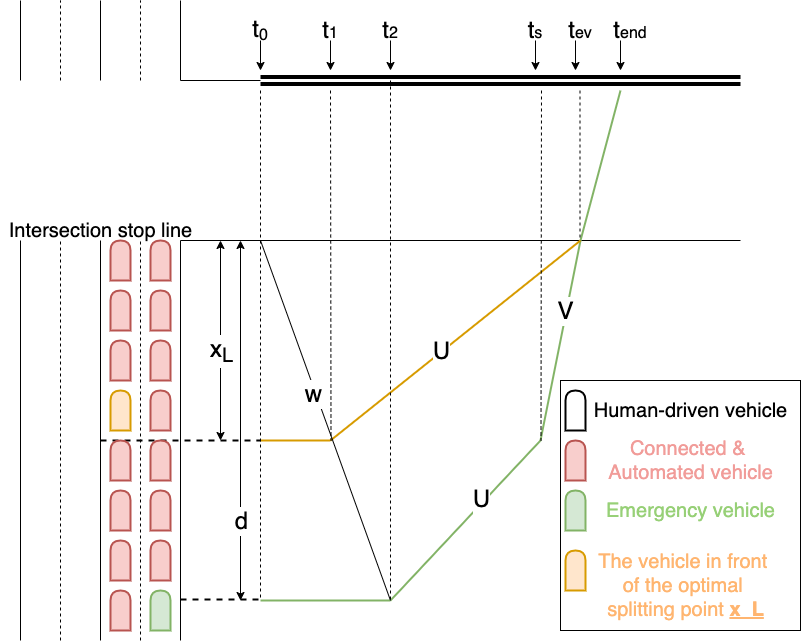}
  \caption{100\% penetration rate of connected vehicles. This scenario is derived from \cite{jordan2013path}. }
  \label{singleInter_highpenetration}
\end{subfigure}
\begin{subfigure}{0.44\textwidth}
  \centering
  \includegraphics[width=\textwidth]{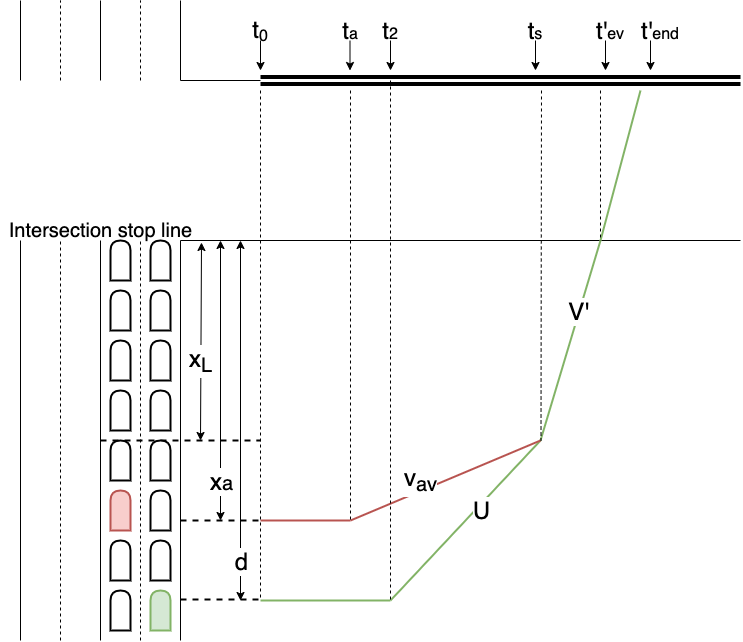}
  \caption{Only one connected automated vehicle is capable of exchanging messages with the EMS vehicle.}
  \label{fig:singleInter_lowpenetration}
\end{subfigure}
\caption{CAV-enabled corridor clearance for emergency vehicles on a congested intersection.}
\label{fig:singleInter}
\end{figure*}

\vj{As an illustrative example, consider the two scenarios presented in Fig.~\ref{fig:singleInter}}. 
If we assume a 100\% CAV penetration rate as in~\cite{jordan2013path}, then the CAV platoon on the left lane (represented as red entities in Fig.~\ref{singleInter_highpenetration}) can split at a given point $x_L$ such that the EMS vehicle (represented as green entities in Fig.~\ref{singleInter_highpenetration}) on the right lane can switch to the left and travel at its desired speed until it reaches the intersection. The idea behind the platoon splitting is to ``sacrifice'' the speeds and travel times of some vehicles in the left lane to assist the maneuvers of the EMS vehicle, which needs to pass the intersection quickly. 

Since it is impossible to achieve a 100\% market penetration rate during the early stages of CAV deployment, all CAVs will not behave exactly the same way as expected. We relax this assumption made in previous work and explore a similar but more realistic scenario where only one CAV can communicate with the EMS vehicle through wireless channels, as shown in Fig.~\ref{fig:singleInter_lowpenetration}. 

\vj{We note that the choice of a single CAV is intentionally made to model a controlability wise hard environment from a system control point of view. A single CAV is the least amount of system control we can introduce (100\% penetration is the highest). However, as we note in the later sections, the introduction of a single CAV instead of multiple CAVs poses different challenges from the control synthesis point of view, which falls beyond the scope of this work, and warrants future work. }

\vj{Moreover, consider the following modeling choices, which are done to emulate short-term deployment conditions and requirements in the real world.}

\begin{itemize}
    \item All traffic signals are operated based on fixed time control rules. 
    \item Neither the traffic signals nor CAVs support vehicle-to-infrastructure communication. This implies that the traffic signals do not support traffic signal preemption by emergency vehicles for queue discharge near intersections~\cite{obrusnik2020queue}. Also, the CAVs do not receive SPaT messages from the traffic signals. 
    \item The CAV receives information regarding the positions and status of the leading and the following vehicle via onboard sensors, as well as the position and speed of the EMS vehicle, through data exchange in real time.
\end{itemize}

Assuming that the time when the traffic signal turns green is $t_0$ and when the EMS vehicle enters the new road segment after it has just crossed the intersection is $t^{'}_{end}$, as shown in Fig.~\ref{fig:singleInter_lowpenetration}. We are interested in finding the EMS vehicle's travel time given in Eq.~\ref{eq:traveltime_ev}. 

\begin{equation}\label{eq:traveltime_ev}
T_{ev} = t^{'}_{end} - t_0 
\end{equation}

Additionally, we are also interested in the travel times of the CAV as seen in Eq.~\ref{eq:traveltime_CAV}. Here, we use the travel time of the CAV as a \textit{proxy} for the travel time of traffic following the CAV, as it is a strict upper bound.
\begin{equation}\label{eq:traveltime_CAV}
T_{cav} = t^{cav}_{end} - t_0 
\end{equation}


\vj{Our goal is to synthesize CAV control strategies that can achieve a balance between reducing the disruption to normal traffic by EMS vehicle maneuvers while also enabling efficient maneuvers for the EMS vehicle.  In light of this, the throughput at the intersection is also an important metric to consider. We denote the throughput as seen in Eq.~\ref{eq:throughtput}, where $N$ denotes the number of vehicles that are able to cross the intersection after the EMS vehicle disruption within the same green phase, and $h_i$ denotes the headway of the $i$th vehicle.}

\begin{equation}\label{eq:throughtput}
q_{inter} = \frac{N}{\sum h_i}
\end{equation}

\vj{Our overall goal, therefore, is to minimize the travel time of both EMS vehicles and CAV (Equation~\ref{eq:traveltime_ev} and Equation~\ref{eq:traveltime_CAV}) while maximizing the throughput at the intersection (Equation~\ref{eq:throughtput}) under short term deployment requirements and conditions. } 



\subsection{Optimal Control Problem}
\label{optimal-control}
\vj{ 
Below, we introduce the formal high-level optimal control problem for EMS corridor clearance problem. }

\vj{ Given vehicle dynamics models $f_{cav}$, $f_{ev}$, and $f_{human}$ for CAV, emergency vehicle, and human vehicle, respectively, we formulate the optimal control problem for EMS corridor clearance problem as follows. As mentioned earlier, our objectives here are threefold. First, we aim to reduce the CAV travel time $T_{ev}$. Second, we also aim to reduce the rest of the traffic travel time as well. This includes reducing CAV travel time $T_{cav}$ and of the rest of the human vehicles $T_{human}^i$ for all human-driven vehicles $i$. Finally, we aim to increase the intersection throughput as defined in Equation~\ref{eq:throughtput}. }

\begin{equation}\label{eq:obj_model-based}
\min J = T_{ev} + T_{cav} + \sum_{i=1}^{n} T_{human}^i - q_{inter}
\end{equation}

such that for every vehicle $i$, 

\begin{equation}
a_i(t) =
\begin{cases}
f_{cav}(h_i(t), \dot{h}_i(t), v_i(t)) & \\ \quad \quad \quad \quad \quad \quad \text{if vehicle $i$ is the CAV}  \\
f_{ev}(h_i(t), \dot{h}_i(t), v_i(t)) & \\  \quad \quad \quad \quad \quad \quad \text{if vehicle $i$ is the EMS vehicle} \\
f_{human}(h_i(t), \dot{h}_i(t), v_i(t)) & \\  \quad \quad \quad \quad \quad \quad \text{otherwise} 
\end{cases}
\end{equation}

\begin{equation}
\begin{aligned}
& \int_{0}^{T_{cav}} v_i(t) dt=x_a, && \text{if vehicle $i$ is the CAV}, \\
& \int_{0}^{T_{ev}} v_i(t) dt=d, && \text{if vehicle $i$ is the EMS vehicle}, \\
& \int_{0}^{T_{human}^i} v_i(t) dt=l_i, && \text{if vehicle $i$ is human driven}.
\end{aligned}
\end{equation}

\begin{subequations}\label{eq:optimal_condition}
\begin{gather}
h_{min} \leq h_i(t) \leq h_{max} \quad \forall t \in [0, T]\\
v_{min} \leq v_i(t) \leq v_{max} \quad \forall t \in [0, T]\\
a_{min} \leq a_i(t) \leq a_{max} \quad \forall t \in [0, T]
\end{gather}
\end{subequations}

\vj{ 
Here $T= T_{ev}$ if the $i$th vehicle is the EMS vehicle, $T= T_{cav}$ if the vehicle is the CAV, and $T= T_{human}^i$ if its the $i$th human-driven vehicle. The maximum and minimum headway is denoted by $h_{min}$ and $h_{max}$. The maximum and minimum velocity is denoted by $v_{min}$ and $v_{max}$. Finally, $a_{min}$ and $a_{max}$ denote the minimum and maximum acceleration limits.} 

\section{Learning for EMS corridor clearance problem}

\vj{Once we model the EMS corridor clearance problem with short-term deployment requirements, the next step is to devise control methods for CAVs in assisting such corridor clearances. This means solving the optimal control problem presented in Section~\ref{optimal-control}. However, it is worth noting that existing research in this field often assumes full control over the system with 100\% penetration of CAVs, access to SPaT messages, and a vehicle dynamics model for planning purposes. These assumptions severely limit the scalability of the proposed methods when dealing with a larger number of control scenarios that would need to be tackled when these controllers are deployed. }

\vj{The reliance on assumptions concerning penetration levels and SPaT messages undermines the robustness of the proposed methods in real-world short-term scenarios. Due to such task underspecifications~\cite{jayawardana2022impact}, one can anticipate substantial performance drops in previously proposed methods when they are implemented. Furthermore, the assumption of vehicle dynamics models necessitates the re-formulation of model-based approaches for each unique scenario, leading to a highly unscalable algorithm design process.}

\vj{In addressing these limitations, we resort to model-free deep reinforcement learning. Model-free DRL does not assume a model of vehicle dynamics and can accommodate imperfections in the environments, and can come up with control strategies that perform well in expectation. Similar strategies have been leveraged for vehicular control with goals such as eco-driving~\cite{jayawardana2022learning} and traffic smoothing~\cite{lichtle2022deploying}. In the following section, we first describe in detail why model-based methods may be sub-optimal for control synthesis in EMS path clearance problems and describe our model-free DRL approach in detail as an alternative.  }

\subsection{Limitations of model-based control for EMS corridor clearance problem}


Designing model-based control strategies for addressing the EMS corridor clearance problem with short-term deployment requirements is a complex task due to several factors. The intricate nature of vehicle dynamics and inter-vehicle dynamics adds complexity, as these interactions are challenging to model accurately. Uncertainties in queue dissipation rates further complicate the design process, as predicting queue dissipation relies on various factors and introduces additional uncertainties. Additionally, the lack of support from Vehicle-to-Infrastructure (V2I) communication systems adds to the difficulties, requiring control strategies to rely solely on vehicle-based information. These challenges highlight the intricacies involved in developing effective model-based control strategies for the EMS corridor clearance problem.

\begin{enumerate}[a.]
\item
\textbf{Vehicle dynamics}: Modeling the dynamics of vehicles is a challenging task, as they exhibit complex behaviors that are difficult to represent with closed mathematical formulas~\cite{schramm2014vehicle}. Compounding this complexity is the fact that vehicles frequently interact with one another, necessitating the inclusion of inter-vehicle dynamics in the models. In practice, due to the complexities involved, previous research often resorts to employing simplified dynamics models to facilitate planning and sacrificing potential benefits~\cite{jordan2013path}. However, it is important to acknowledge that even if a more comprehensive dynamics model is utilized, it introduces additional complexities to the optimization problem, making it more challenging to solve even with industry-standard solvers~\cite{altche2017high}.
\item
\textbf{Queue dissipation rates}: Accurate estimation of queue lengths at intersections is crucial for formulating the control problem in model-based methods. However, this task is far from straightforward. One common approach is to utilize fundamental diagrams to estimate queue lengths~\cite{liu2009real}, but this assumes the availability of such diagrams for the specific roadways involved. Moreover, in certain cases, the optimization problem may require queue information not only for the immediate intersection but also for subsequent intersections. This adds another layer of complexity to queue estimation, as it necessitates access to queue lengths at multiple intersections simultaneously. As a result, tackling this challenge poses significant difficulties for model-based control strategies.

\item
\textbf{Traffic signal timing}: An essential aspect of model-based corridor clearance is the availability of real-time SPaT messages from traffic lights~\cite{li2018signal}. However, obtaining this information can be challenging since it involves installing roadside units on traffic signals. In cases where direct access to SPaT data is not feasible, model-based methods must account for the stochastic nature of traffic signals. Consequently, they need to devise control strategies that perform optimally on average. These formulations tend to be complex and demand substantial real-time computing power for effective solutions.

\item
\textbf{The communication between vehicles and infrastructure}: Accurate data, including Signal Phase and Timing (SPaT) messages from traffic signals and queue lengths at intersections, is essential for effective model-based methods in problem formulation. To achieve this, wireless communication is required to transmit such information from the infrastructure to vehicles. However, meeting this requirement poses challenges, primarily due to the higher costs and upfront investment involved in establishing roadside infrastructure for Vehicle-to-Infrastructure (V2I) communication compared to the more feasible Vehicle-to-Vehicle (V2V) communication~\cite{ligo2018cost}.

\end{enumerate}

\subsection{Deep reinforcement learning for EMS corridor clearance}

Due to the limitations outlined earlier in model-based control design, applying model-based control synthesis in complex real-world traffic scenarios poses significant challenges. However, deep reinforcement learning presents a promising alternative, capable of effectively handling these complexities during training and generating high-performing and realistic controllers.

Unlike model-based control, deep reinforcement learning does not rely on a closed-form model of vehicle dynamics. Instead, it leverages arbitrarily complex traffic simulators, making it more adaptable to real-world conditions. Additionally, it has the ability to learn control strategies that perform optimally on average, even in the absence of traffic signal timing data and queue lengths at intersections. Another key advantage of deep reinforcement learning lies in its proficiency in handling imperfect information, surpassing model-based methods in this aspect. Consequently, it reduces the dependency on physical communication channels to transfer information, allowing for a more streamlined and minimalistic communication setup. Therefore, we next present our deep reinforcement learning-based EMS corridor clearance controller.  

\subsubsection{Partially observable markov decision process}

The EMS corridor clearance problem can be formulated as a discrete-time \textit{partially observable Markov Decision Process (POMDP)} by extending the MDP formulation introduced in section~\ref{MDP_basic} with two more components, observation space $\Omega$ and conditional observation probabilities $O:\mathcal{S}\times\Omega  \rightarrow  \Delta(\Omega)$. We characterize the POMDP for the EMS corridor clearance problem as follows:
\begin{itemize}
    \item 
    \textbf{Observations}: An observation consists of the relative distance and the speed of the immediate leading and following vehicles of the CAV, the speed, the lane, and the position of the emergency vehicle and the speed and the position of the CAV. In other words, a state $s \in \mathcal{S}$ is defined as  ($v_{lead},h_{lead},v_{follower},h_{follower},v_{ev},l_{ev},p_{ev},v_{cav},p_{cav} $) where $lead$ and $follower$ are the immediate leading and following vehicles of the CAV. $p_{cav} - p_{ev} < 0$ means that the EMS vehicle is in a downstream location with respect to the CAV, and vice versa. Since there are physical limitations for the maximum possible range of wireless communication between EMS vehicle and the CAV, we set $p_{cav} - p_{ev}\in (-r,r)$ where $r$ is the maximum range of wireless communication between vehicles. We do not assume nor require any form of communication between CAV and the upcoming traffic signals and therefore do not include any state features related to the upcoming traffic signals. Our devised control policies are consequently not dependent on the Vehicle to Infrastructure communications. However, one can possibly develop better-informed control policies for the CAV given access to Signal Phase and Timing (SPaT) messages as we further discuss in the future work of this study. 
    
    \item \textbf{Actions}: The action is the acceleration command $\alpha \in (\alpha_{min}, \alpha_{max})$ for the longitudinal control of the CAV.
    
    \item \textbf{Transition Function}: We do not explicitly define the stochastic transition function. Instead, we use microscopic simulations to sample $s_{t+1} \sim p(s_t, a_t)$ where the simulator applies the accelerations to vehicles and updates the traffic signals. 
    
    \item \textbf{Reward}: The goal of the CAV is twofold. First, it aims to clear the path for the EMS vehicle facilitating the EMS vehicle to travel at its desired speed by performing an emergency lane changing. Second, it aims to reduce the influence of its actions on normal traffic. The first objective requires the CAV to block the traffic behind it to assist the EMS vehicle's lane-changing maneuver. The second objective requires that the regular traffic (e.g., the speeds and travel time of the CAV and other vehicles except for the EMS vehicle) should also be considered in the reward function. Therefore, we design a composite reward function as given in Eq.~\ref{reward}, where $\nu_1,\nu_2,\nu_3$ and $\nu_4$ are a hyper-parameters that balances the trade-off between the EMS vehicle and the impact on regular traffic.

\end{itemize}


\begin{equation}\label{reward}
  \setlength{\arraycolsep}{0pt}
  r(s,a): = \left\{ \begin{array}{ l l }
    \nu_1*v_{ev} + \nu_2*v_{cav}, & \quad \text{if } p_{cav} - p_{ev} \leq 0 \\
    & \quad  \wedge \ l_{ev} \neq l_{cav} \\
    \nu_3*v_{ev} + \nu_4*v_{cav}, & \quad \text{if } p_{cav} - p_{ev} > 0 \\
    v_{cav}, & \quad l_{ev} = l_{cav} \\
    & \quad \vee \textit{ no EMS vehicle}
  \end{array} \right.
\end{equation}

In Eq.~\ref{reward}, the first case corresponds to when the CAV is ahead of the EMS vehicle but in a different lane. In such case, we encourage high EMS vehicle speed more than high CAV speed by setting $\nu_1 > \nu_2$. In the second case, the EMS vehicle is ahead of the CAV. Therefore, we encourage both EMS vehicle speed and the CAV speed by setting $\nu_3 \approx \nu_4$. Finally, the third case corresponds to either when CAV is ahead of the EMS vehicle, but they share the same lane, or only the CAV is present, and no EMS vehicle is present in the system. In either case, we set $v_{cav}$ directly as the reward.

The reason we added $v_3*v_{ev}$ when $p_{cav}-p_{ev} > 0$ AND $l_{ev} = l_{cav}$ is actually for safety purposes. What we found from the experiments is that this “regularization term”, $v_3*v_{ev}$, can reduce rear-end collision potentials between the CAV to the EMS. If we only use the term containing the CAV’s speed $v_4*v_{cav}$ in the reward function, the CAV will have the tendency to start to accelerate too early (and aggressively) right after the moment when the EMS has just finished lane changing, leading to more collision risks.

\section{Experiments}
\label{research-questions}
\vj{In our evaluation of the proposed Deep Reinforcement Learning controller, we adopt a two-fold approach to assess its effectiveness. Firstly, we conduct a qualitative assessment of the complexity involved in designing model-based methods when traffic settings change. We demonstrate the additional effort required to design a model-based controller when transitioning from one intersection to two intersections. Simultaneously, we highlight the ease with which model-free DRL can handle such changes. Second, we perform quantitative assessments of the proposed method by comparing the performance of the DRL controller with that of model-based methods. Below, we present these two evaluation directions as specific research questions that we address in this study.}

\subsection{Research Questions}
\vj{
We evaluate the DRL controller to answer four main research questions: 
\begin{itemize}
    \item \textbf{Q1}: Qualitatively, how difficult is it to design model-based methods and DRL methods for changing traffic settings?
    \item \textbf{Q2}: How much better is the performance of the DRL controller compared to the model-based method, measured based on the EMS vehicle and CAV \textbf{travel times} and the \textbf{throughput} at the congested intersection after the EMS disruption event?
    \item \textbf{Q3}: How suboptimal are the DRL and model-based controllers?
    \item \textbf{Q4}: What are the benefits and drawbacks of model-based and DRL controllers with respect to computation time? 
\end{itemize}
  }

\vj{
In answering Q1, we first extend \cite{jordan2013path} to scenarios other than 100\% CAV penetration as our baseline. Then we further extend it to accommodate a single intersection and two intersections dynamics. Simultaneously, we show how we solve single and two intersection settings with DRL controller. Throughout this analysis, we aim to qualitatively illustrate the challenges associated with model-based control design in contrast to the advantages offered by DRL-based control design.}

\vj{
In answering Q2, we show in Section~\ref{performance} that our DRL controller outperforms the model-based controller by reducing the travel time of the CAV while keeping the travel time of the EMS vehicle as same as the model-based controller, which aligns well with our goals. Furthermore, we show that the DRL controller achieves better throughput compared to model-based controller indicating its ability to trade-off between EMS vehicle travel time and impact on regular traffic.}

\vj{
Our analysis of Q3 indicates that the DRL controller is closer and sometimes better than the theoretically optimal controller developed by using graphical analysis and shockwave theory. We provide further explanation of this phenomenon in Section~\ref{optimal-model}.}

\vj{Finally, in answering Q4, we provide a comparative analysis of the two controllers concerning their computational requirements. While it is true that model-based methods may exhibit computational efficiency due to the absence of training phases, we must take into account the time required for problem formulation. In certain cases, despite their computational advantages, model-based methods might not necessarily be the optimal choice when considering the overall time needed for problem setup and execution. }

\section{Experimental Setup}

\subsection{Network and simulation settings}
We use SUMO microscopic traffic simulator~\cite{SUMO2018} for our experimental simulations. \vj{We model a $1 \times 1$ and $1 \times 2$ grids consisting of one and two 4-way intersections with through traffic only}. In each intersection, each incoming and outgoing approach has two lanes 175m long. The speed limit for non-EMS vehicles is $15 m/s$ while $35 m/s$ for the EMS vehicle. All non-CAV vehicles are controlled by Intelligent Driver Model (IDM)~\cite{Treiber2000CongestedTS}. All vehicles are subjected to a maximum permitted acceleration of $3m/s^{2}$ and a maximum permitted deceleration of $3m/s^2$. Each intersection is controlled by a fixed time signal plan in which each 31s-long green phase is followed by a 6s-long yellow phase before turning red. We use a uniform vehicle inflow rate of 1000 vehicles/hour to simulate vehicle arrivals. 

\subsection{Training hyperparameters}

We model our experimental simulations in the Flow framework~\cite{wu2021flow}. A neural network of three hidden layers of 32 neurons each is used as our controller $\pi_{\theta}$ and trained with the Proximal Policy Optimization (PPO) algorithm~\cite{Schulman2017ProximalPO}. All training parameters are initialized to the defaults specified by the Flow framework. The controller is trained using 0.2M episodes with ten parallel workers. We use a simulation step duration of 0.5 seconds and a horizon of 600 steps. A discounting factor of 0.999 and a learning rate of 0.001 were used in the experiments.

\section{Simulation results and discussion}
\label{results}

\vj{In this section, we provide detailed analysis to answer the four research questions presented in Section~\ref{research-questions}.}

\subsection{Ease of problem formulation}

\vj{In this subsection, we undertake a qualitative assessment to address Q1, comparing the challenges involved in designing model-based controllers versus DRL controllers for the corridor clearance problem. Due to space constraints, we provide a walkthrough of the model-based method's controller design for both single and two intersections in the Appendix detailing the necessary modifications needed to extend the single intersection model-based controller to a two-intersection model-based controller. Thus, in this section, we present the changes required in DRL controller design when extending from single intersection to two intersections.}

\vj{As highlighted in the appendix, the process of designing model-based controllers for adapting to changing traffic settings proves to be time-consuming and unscalable, necessitating complete reformulations from scratch. Conversely, the DRL controller design offers a contrasting advantage by providing a straightforward approach to accommodate such changes. Notably, we demonstrate that the only modification needed in the DRL controller design lies in adjusting the reward coefficients. For reference, we present the reward coefficients utilized in the two scenarios in Table~\ref{table:hyper-parameters-1} and \ref{table:hyper-parameters-2}.}

\vj{By requiring changes solely in the reward coefficients, the fundamental structure of the Markov Decision Process remains unaffected. This results in significant savings in controller design time and enhances practical scalability, making the DRL approach a more attractive and efficient option for handling varying traffic scenarios. }

\vj{One of the limitations of DRL controller design lies in the process of finding the optimal coefficients. Although we employ grid search to explore various configurations and identify the best ones, this step can be computationally expensive. It is worth noting that discovering optimal hyperparameters in reinforcement learning remains an active research area and extends beyond the scope of this study.}

\vj{Nevertheless, we anticipate that the adoption of advanced hyperparameter optimization techniques, such as Population-based Training~\cite{jaderberg2017population}, can offer potential solutions to this challenge. These techniques have shown promise in efficiently navigating the hyperparameter space and could alleviate the issue of finding the right coefficients more effectively.}

\begin{figure}[tb!]
\centering
\includegraphics[width=0.5\textwidth]{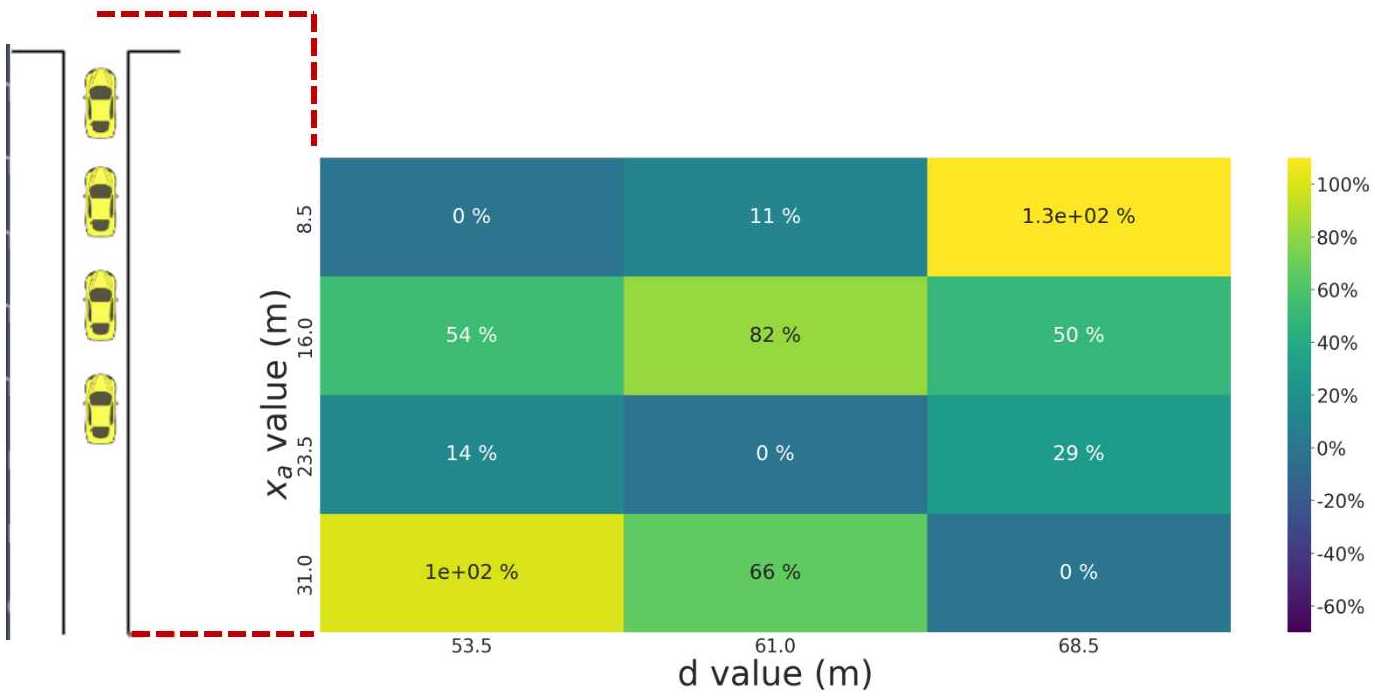}
\caption{Percentage throughput difference under the DRL controller and the model-based controller with varying $x_a$ and $d$ values for the single-intersection scenario. Positive values indicate throughput under DRL controller is higher than that of model-based controller. }
\label{fig:throughput_single}
\end{figure}

\begin{figure}[tb!]
\centering
\includegraphics[width=0.5\textwidth]{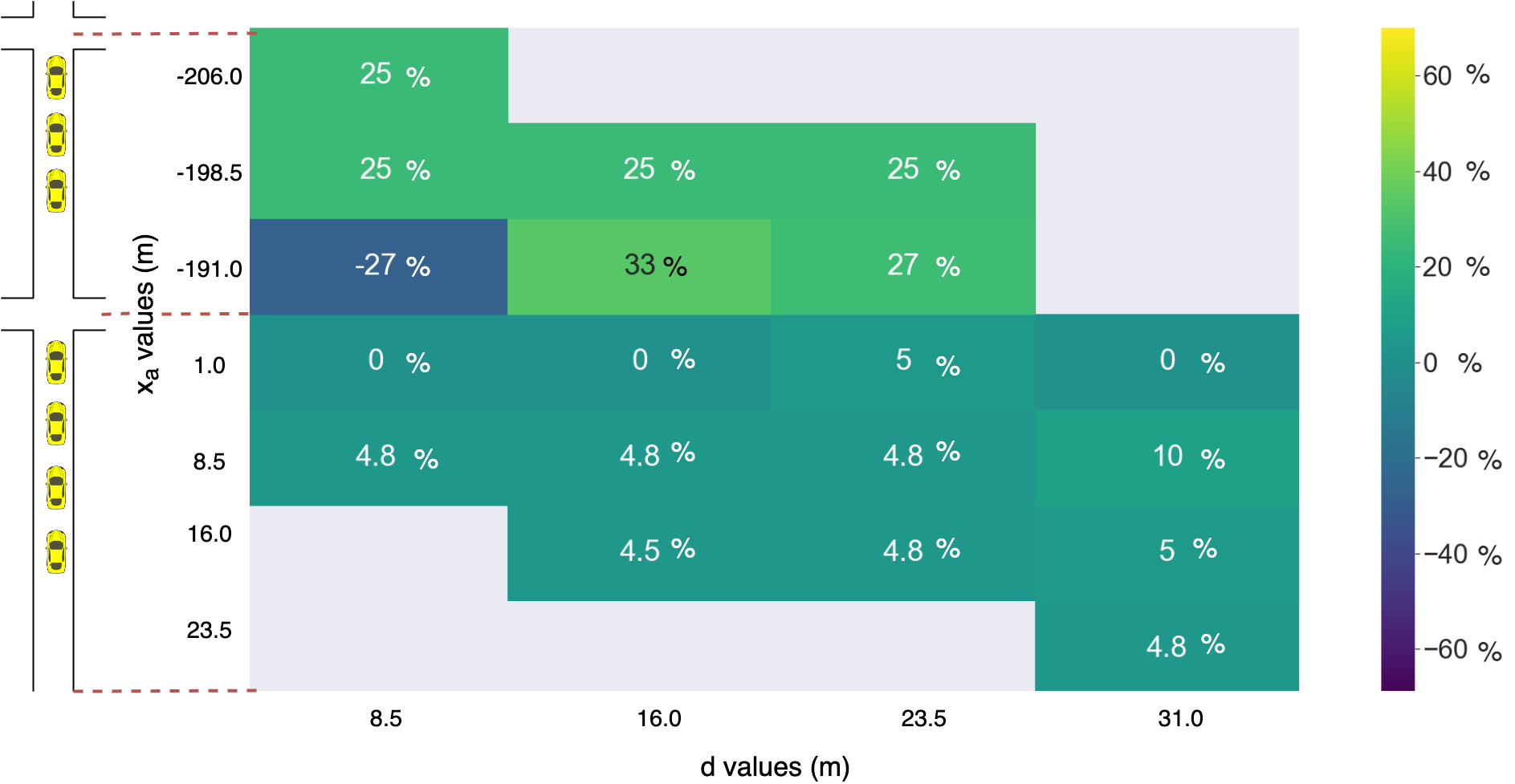}
\caption{Percentage throughput difference under the DRL controller and the model-based controller with varying $x_a$ and $d$ values for the two-intersection scenario. Positive values indicate throughput under DRL controller is higher than that of model-based controller. }
\label{fig:throughput}
\end{figure}

\begin{figure*}[h]
\centering
\includegraphics[width=0.85\textwidth]{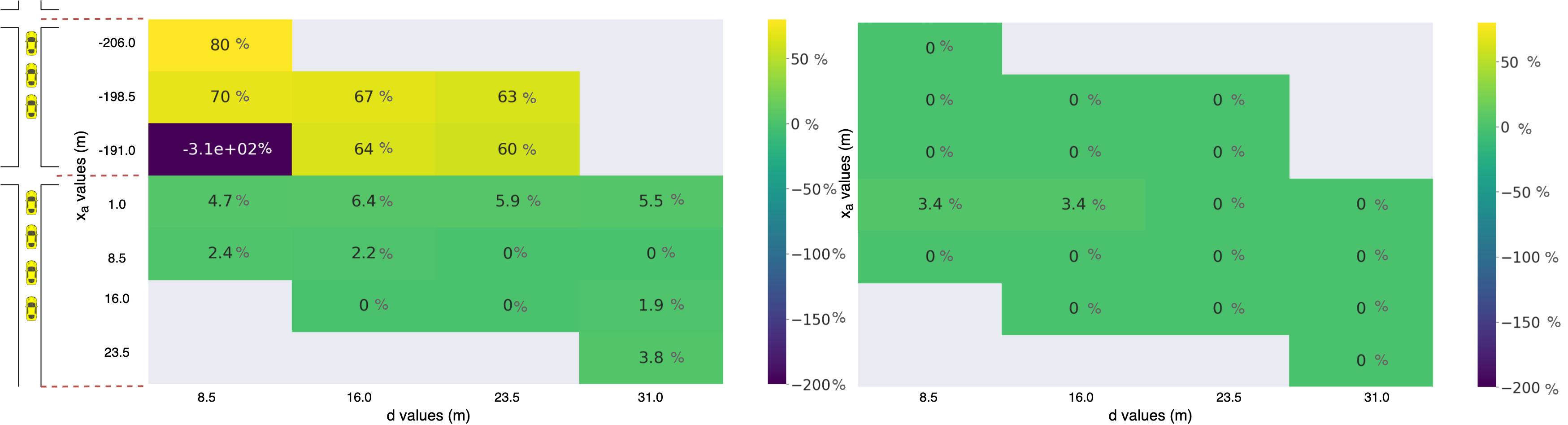}
\caption{Percentage travel time difference of the CAV (left) and the EMS vehicle (right) under the DRL controller and the model-based controller with varying $x_a$ and $d$ values. Positive values indicate CAV or EMS vehicle travel time under DRL controller is smaller than that of model-based controller. }
\label{fig:performacne}
\end{figure*}

\begin{table}[t]
\centering
    \caption{Reward hyper-parameter configuration for single intersection scenario}
		\begin{tabular}{c|c||c|c}
			\hline
            Parameter & Value & Parameter & Value \\
            \hline
            ${\nu_1}$ & 0.1 & ${\nu_2}$ & 0.9 \\ 
            \hline
            ${\nu_3}$ & 0 & ${\nu_4}$ & 1 \\ 
            \hline
		 \end{tabular}
		\label{table:hyper-parameters-1}
\end{table}

\begin{table}[t]
\centering
    \caption{Reward hyper-parameter configuration for two intersections scenario}
		\begin{tabular}{c|c||c|c}
			\hline
            Parameter & Value & Parameter & Value \\
            \hline
            ${\nu_1}$ & 1 & ${\nu_2}$ & 0.6 \\ 
            \hline
            ${\nu_3}$ & 1 & ${\nu_4}$ & 1 \\ 
            \hline
		 \end{tabular}
		\label{table:hyper-parameters-2}
\end{table}

\subsection{Performance comparison}
\label{performance}


In answering Q2, we measure the throughput difference between the DRL and the model-based controllers. Furthermore, we measure the travel times of the CAV and the EMS vehicles collected from simulation experiments. Combining the two metrics allows us to quantify the trade-off between EMS vehicle travel time and its impact on traffic flow. 

On the one hand, results from the single-intersection (Fig.~\ref{fig:throughput_single} and two-intersection simulations (Fig~\ref{fig:throughput}) indicate that the DRL controller can maintain superior throughput in the majority of the cases.  

On the other hand, we observe the DRL controller can significantly reduce the travel time of the CAV in most cases while keeping the EMS vehicle travel time as same as the model-based controller. This observation is highlighted by the percentage difference in CAV travel times under both controllers, as shown in Figure~\ref{fig:performacne}.

When considering potential near-term implementation, these findings collectively suggest that even a solitary DRL can improve overall traffic efficiency (i.e., ensuring high throughput) without compromising on the travel time of EMS vehicles.

In all figures, the two axes $x_a$ and $d$ represent different traffic settings, in which $x_a$ denotes the actual position of the CAV and $d$ denotes the position of the EMS vehicle in the queue before the queues start to discharge. For given values of $x_a$ and $d$, a cell with a positive percentage means that the CAV governed by the DRL controller achieves less travel time or higher throughput than the model-based controller for that particular setting. 

\begin{figure*}[bt!]
\centering
\begin{subfigure}{0.48\linewidth}
  \centering
  \includegraphics[width=\linewidth]{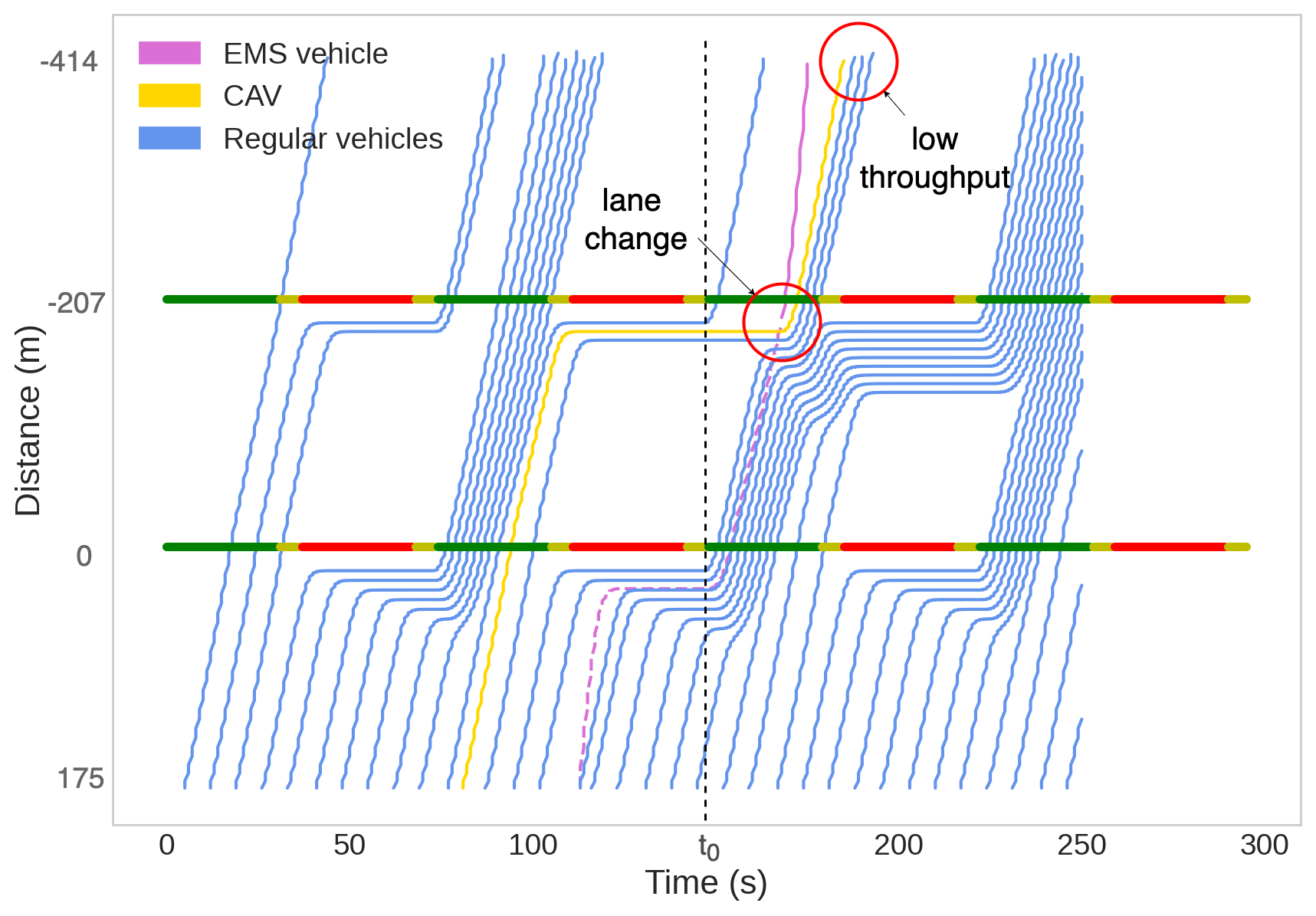}
  \caption{Time-space diagram of model-based control vehicle trajectories}
  \label{fig3a}
\end{subfigure}
\begin{subfigure}{0.48\linewidth}
  \centering
  \includegraphics[width=\linewidth]{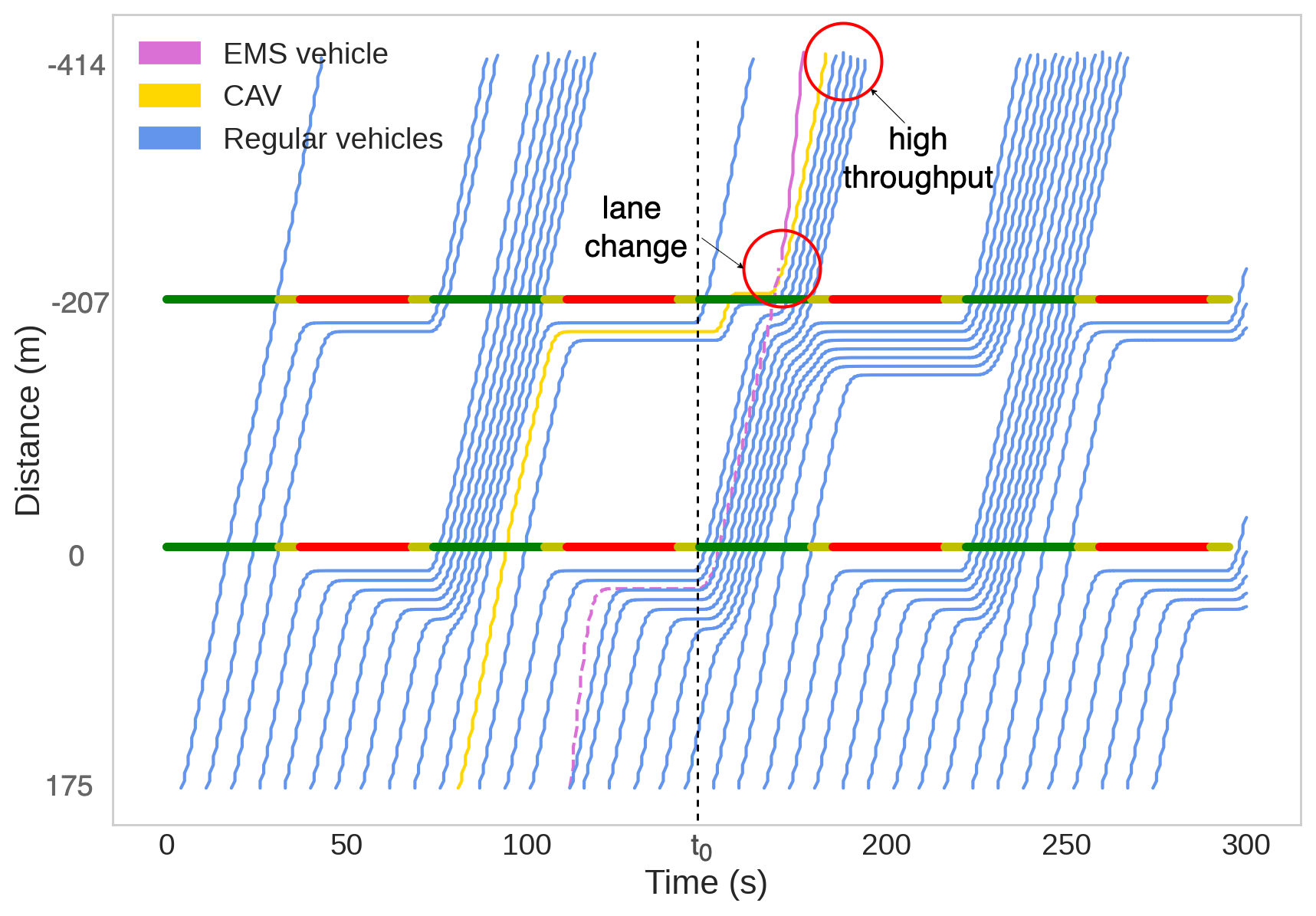}
  \caption{Time-space diagram of DRL controller-based vehicle trajectories}
  \label{fig3b}
\end{subfigure}
\caption{Time space diagrams illustrating a representative instance of the behaviors of the yellow region in Figure~\ref{fig:performacne}. Only the vehicles on the left lanes are considered with the exception of the EMS vehicle where pink dash-line indicates its movement on the right lane. Under the model-based controller, CAV stops on the incoming approach of the second intersection (Figure~\ref{fig3a}) causing all of the following vehicles of the CAV to stop or decelerate producing a lower throughput. Under the DRL controller (Figure~\ref{fig3b}), CAV demonstrates an intelligent behavior by creating the lane change opportunity for the EMS vehicle after crossing the intersection facilitating significantly less perturbation to the traffic flow, and high throughput.}
\label{ts1}
\end{figure*}

\begin{figure*}[bt!]
\centering
\begin{subfigure}{0.48\linewidth}
  \centering
  \includegraphics[width=\linewidth]{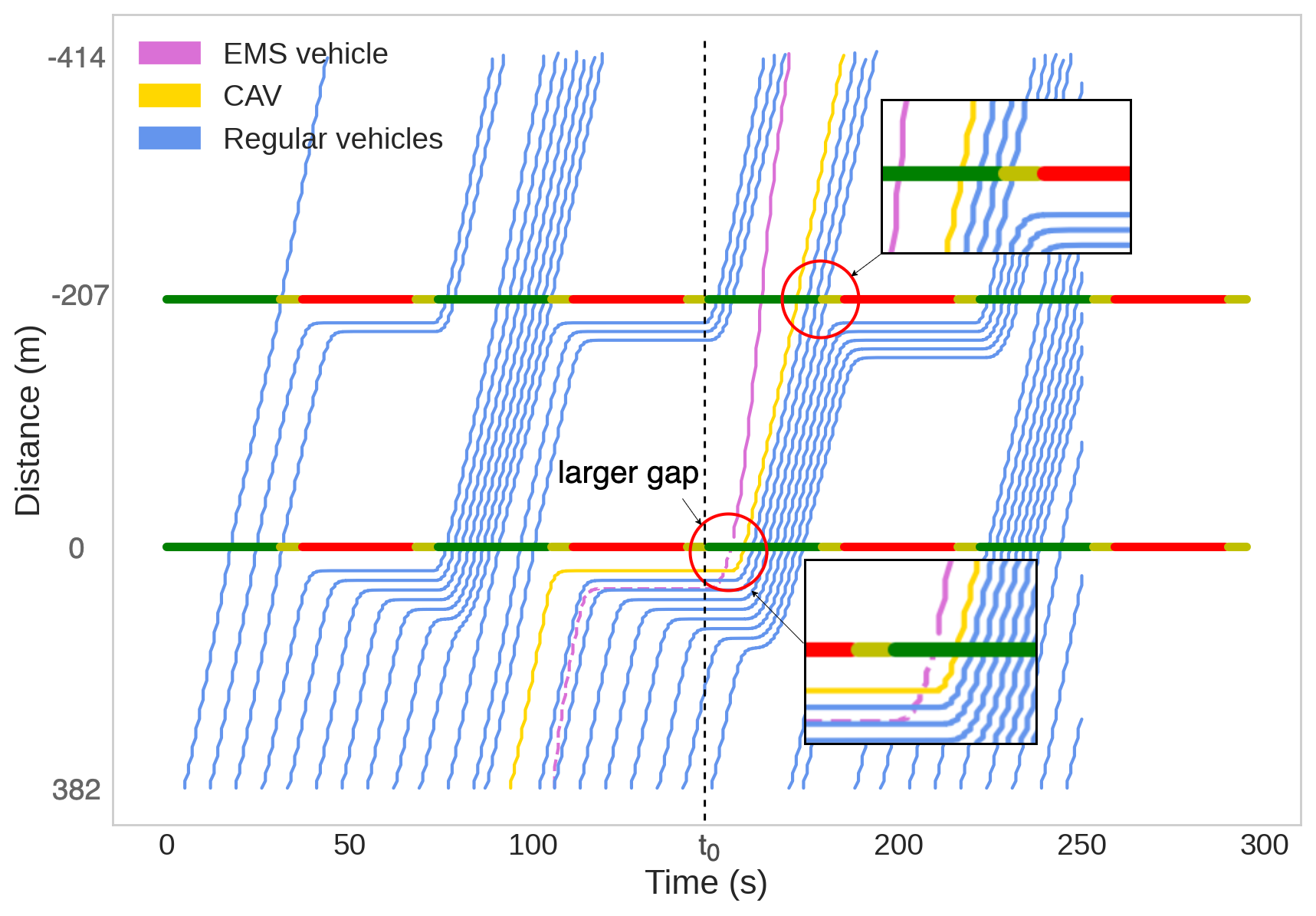}
  \caption{Time-space diagram of model-based control vehicle trajectories}
  \label{fig11a}
\end{subfigure}
\begin{subfigure}{0.48\linewidth}
  \centering
  \includegraphics[width=\linewidth]{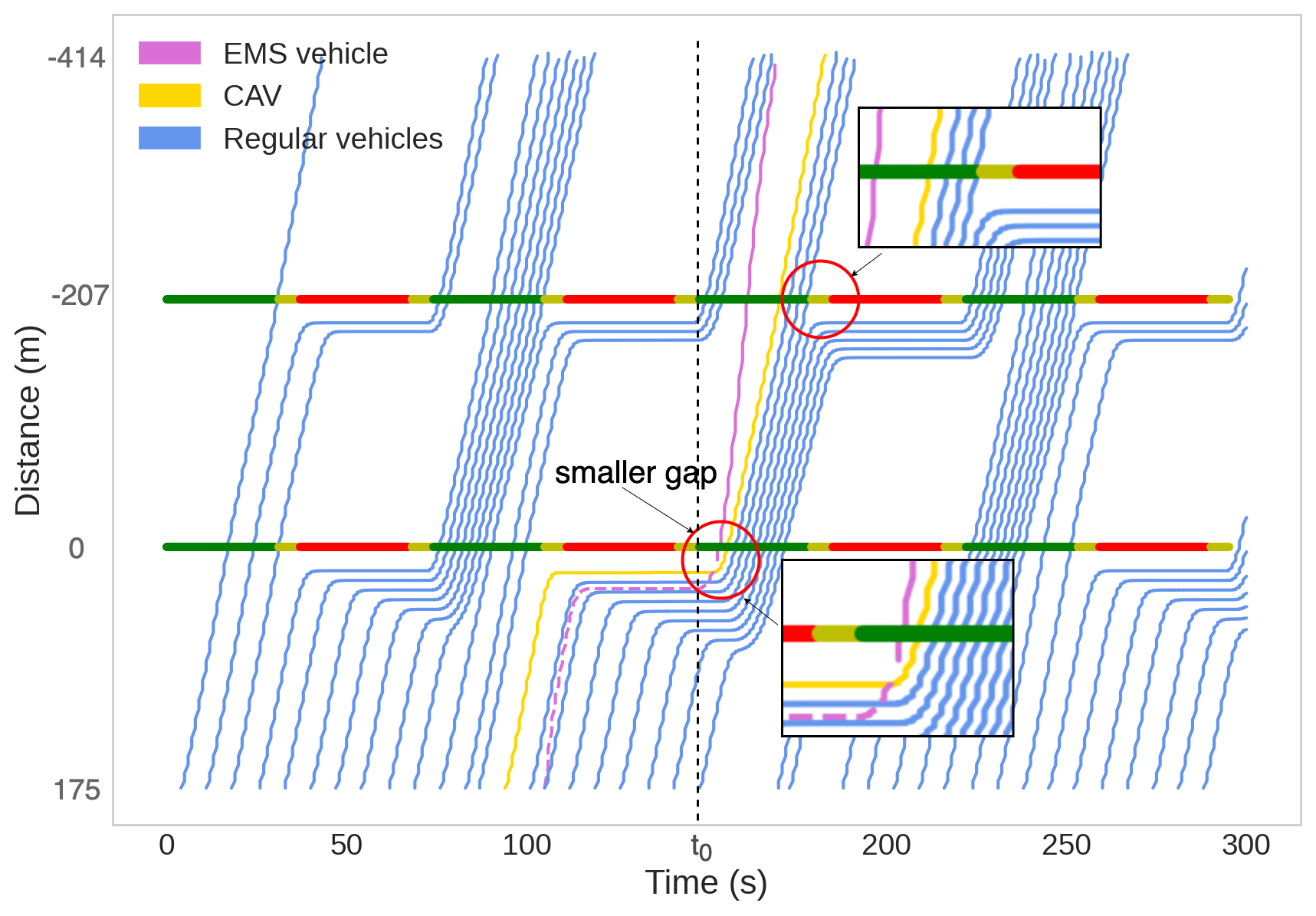}
  \caption{Time-space diagram of DRL controller based vehicle trajectories}
  \label{fig11b}
\end{subfigure}
\caption{Time space diagrams illustrating a representative instance of the behaviors of the green region in Figure~\ref{fig:performacne}. Only the vehicles on the left lanes are considered with the exception of the EMS vehicle where the pink dash line indicates its movement on the right lane. Under the model-based controller, CAV demonstrates a conservative behavior (Figure~\ref{fig11a}) while under DRL controller, CAV demonstrates a proactive behavior (Figure~\ref{fig11b}). The proactive behavior of DRL controller makes a safe opportunity for the following vehicles unlike in the model-based controller where some of the following vehicles cruise through the intersection during the yellow light, making them vulnerable to a stop.}
\label{ts2}
\end{figure*}

\begin{figure*}[bt!]
\centering
\begin{subfigure}{0.48\linewidth}
  \centering
  \includegraphics[width=\linewidth]{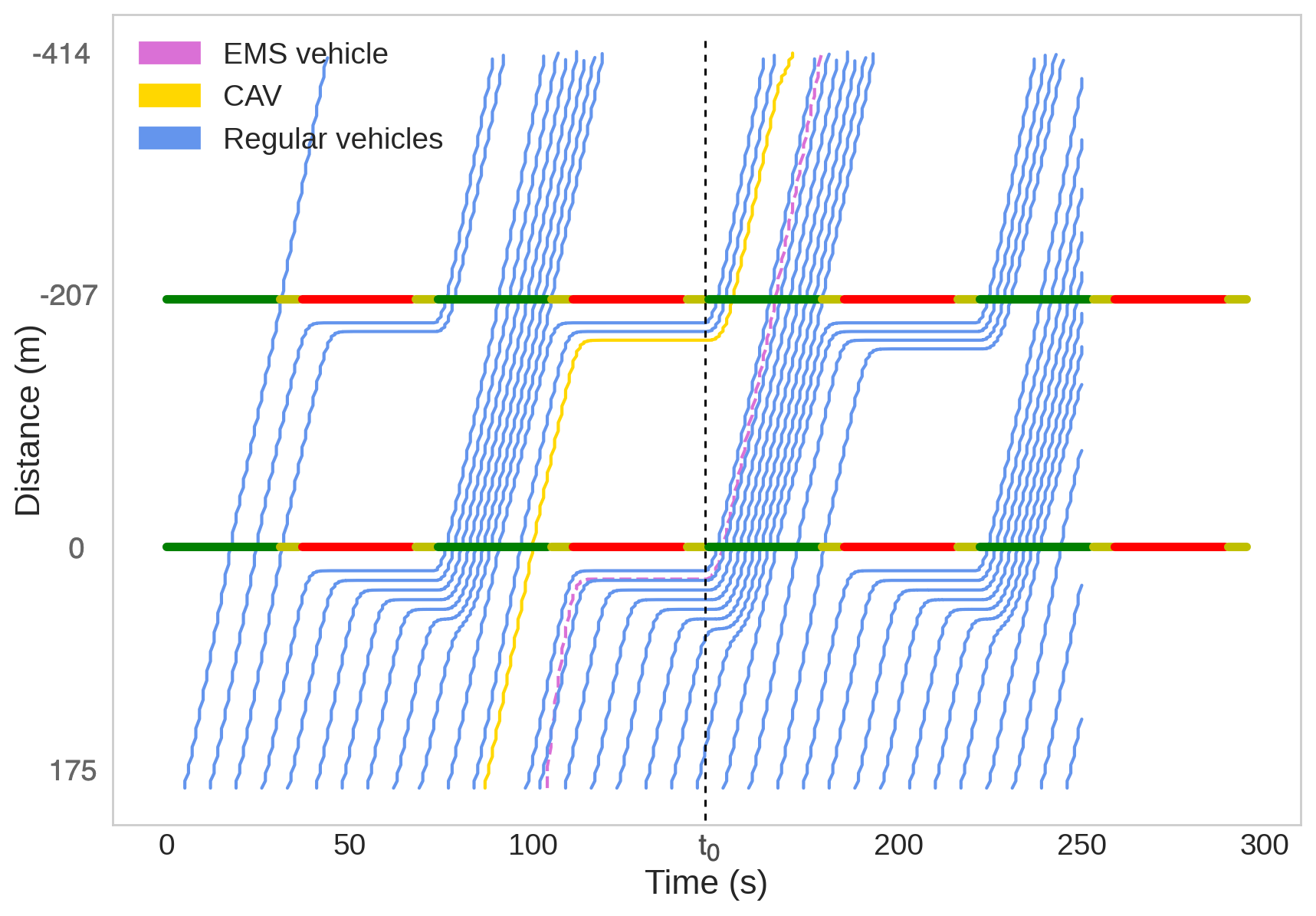}
  \caption{Time-space diagram of model-based controller based vehicle trajectories}
  \label{fig4a}
\end{subfigure}
\begin{subfigure}{0.48\linewidth}
  \centering
  \includegraphics[width=\linewidth]{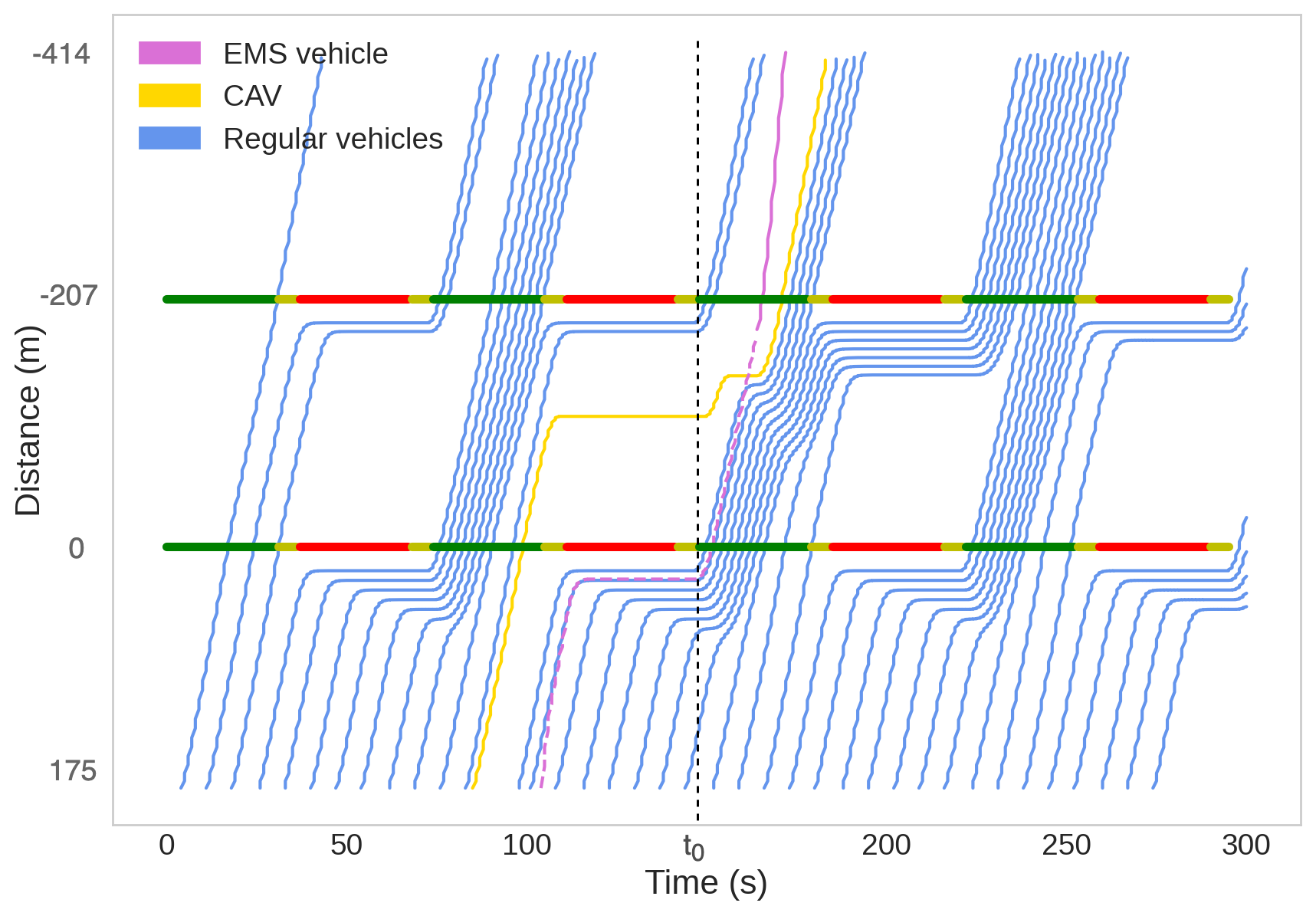}
  \caption{Time-space diagram of DRL controller based vehicle trajectories}
  \label{fig4b}
\end{subfigure}
\caption{Time space diagrams illustrating the behaviors of the purple region in Figure~\ref{fig:performacne} ($x_a=-191.0$ and $d=8.5$) under the model-based and DRL controllers. Only the vehicles on the left lanes are considered with the exception of the EMS vehicle where pink dash-line indicates its movement on the right lane. While CAV under the model-based controller does not wait for the EMS vehicle, DRL controller makes the CAV waits causing increase in travel time of the CAV. 
  } 
\label{ts3}
\end{figure*}

In highlighting the proficiency of the DRL in balancing the travel time of the EMS vehicle and its effect on general traffic, we showcase three distinct behaviors exhibited by the DRL controller, setting it apart from the model-based controller. These three types of behaviors correspond to the three colored regions in Figure~\ref{fig:performacne}a: green, purple, and yellow. To illustrate further on these behaviors, we select a representative instance of each of the behaviors and plot the respective time-space diagrams as in Figure~\ref{ts1} (yellow), Figure~\ref{ts2} (green), and Figure~\ref{ts3} (purple).  
In Figure~\ref{ts1}, we illustrate the vehicle behaviors in the yellow region of Figure~\ref{fig:performacne}a. We recognize this scenario corresponds to the class of scenarios when the CAV stops at the second intersection and the EMS vehicle stops at the first intersection (negative $x_a$ values and positive $d$ values). Under this setting, the DRL controller-based CAV achieves a significant reduction in travel time with the same EMS vehicle travel time as in the model-based controller while producing a 25\% improvement in throughput. Under the model-based controller, CAV stops on the incoming approach of the second intersection, creating an opportunity for the EMS vehicle to lane change and attain a higher speed. However, while minimizing the EMS travel time, this behavior by the CAV causes all of the following vehicles of the CAV to stop or decelerate, producing a lower throughput. In contrast, under the DRL controller (Figure~\ref{fig3b}), CAV demonstrates a more intelligent behavior in balancing the needs between reducing EMS travel time and minimizing the impact on traffic flow. Specifically, the CAV creates a lane-changing opportunity for the EMS vehicle after crossing the intersection rather than before the intersection, as the model-based controller does. This behavior of the CAV significantly reduces the disruption to the traffic flow and generates a higher throughput during the green phase of the traffic signals.

Figure~\ref{ts2} corresponds to a representative scenario of the green region of Figure~\ref{fig:performacne}a. We recognize they depict scenarios where both vehicles stop at the first intersection due to red light (positive $x_a$ values and positive $d$ values). The selected instance corresponds to $x_a = 1m$ and $d = 16m$. Under this setting, the learned DRL controller reduces the travel time of the CAV by 6.4\%. By observing both CAV and EMS vehicle behaviors in Figure~\ref{ts2}, we see that the CAV demonstrates a more conservative movement under the model-based controller in waiting for the EMS vehicle to lane change in front of it. In particular, it waits until a larger gap is created between itself and the EMS vehicle before accelerating. In contrast, DRL controller-based CAV acts more proactively and accelerates as a sufficient gap between the CAV and the EMS vehicle is created. This proactive behavior enables the CAV to obtain reduced travel time compared to the model-based controller, which explains the results shown in Figure~\ref{ts2}. Moreover, the proactive behavior of the DRL controller-based CAV creates a more safe opportunity for the vehicles following the CAV to escape the red light in the second intersection. In contrast, under the model-based controller, some of the following vehicles cruise through the intersection during the yellow light, making them vulnerable to a potential stop (and hence reducing throughput).

Finally, in Figure~\ref{ts3} we show the behavior in the purple region of Figure~\ref{fig:performacne}a. We recognize this instance as the only failure case of the DRL controller when compared with the model-based controller. It depicts a scenario in which the model-based controller is acting according to the special condition specified by Equation~\ref{eq:optimal_condition}. This causes the CAV under model-based controller to not to assist the EMS vehicle and therefore drive as usual. In contrast, the CAV under the DRL controller waits in the hope of assisting the EMS vehicle. While such behavior makes the travel time of the EMS vehicle marginally better, the disruption it causes on the CAV and the rest of the following traffic is significant as can be seen from the -27\% throughput difference. However, this particular instance belongs to the class of instances where the CAV stops at the second intersection, and the EMS vehicle stops at the first intersection (negative $x_a$ values and positive $d$ values). As we explained earlier with Figure~\ref{ts1}, all other instances in this class show a significantly better performance. Therefore, we believe this specific behavior in this instance may occur due to a lack of exploration in training which causes the DRL agent to underperform.

The reasons for the different performance between the model-based and the DRL controllers lie in two aspects. First, the model-based controller is based on a static assumption of $w$, the shockwave speed of the queue discharging rate. In other words, whether the model-based controller can issue ``correct'' commands for the longitudinal control of the CAV depends on how close the estimated value derived from historical traffic data is close to the actual value of $w$. This task is challenging and error-prone if real-time traffic settings diverge from historical data. In our experiments for the model-based controller, we use $w$ = 10, which achieves the best performance in the EMS travel times after fine-tuning with multiple rounds of simulations.

Second, we have consistently observed in our simulation that the CAV controlled by the DRL controller appears to be more \textit{agile} and \textit{intelligent}. It is \textit{agile} in the sense that it causes the CAV to start to move earlier than the model-based controller after the EMS vehicle has switched to the left lane, which can help the traffic flow to recover sooner after being disrupted by the EMS vehicle maneuvers (Figure~\ref{ts2}). It is \textit{intelligent} in the sense that it selects a less disruptive splitting point for lane changing by factoring in the rest of the traffic (Figure~\ref{ts1}). Such behaviors are consistent with our objectives of designing a DRL controller for minimizing the impact on traffic flow while maintaining the travel speed of the EMS vehicle at the desired level.

\begin{figure*}[tb!]
\centering
\includegraphics[width=0.85\textwidth]{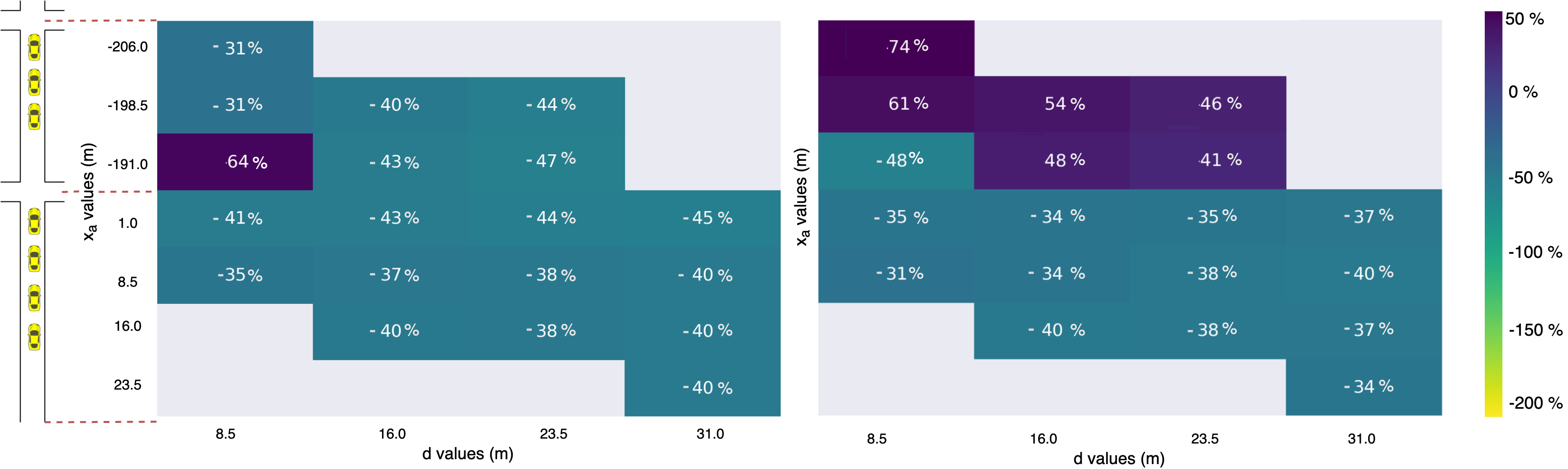}
\caption{Percentage travel time difference of the CAV under the model-based controller (left) and the DRL controller (right) compared against the oracle. Results are reported with varying $x_a$ and $d$ values. Positive values indicate CAV travel time under the DRL or model-based controller is smaller than that of oracle while negative values indicate CAV travel time under the DRL or model-based controller is greater than that of oracle (higher the better).}
\label{fig:cav-opt}
\end{figure*}

\begin{figure*}[tb!]
\centering
\includegraphics[width=0.85\textwidth]{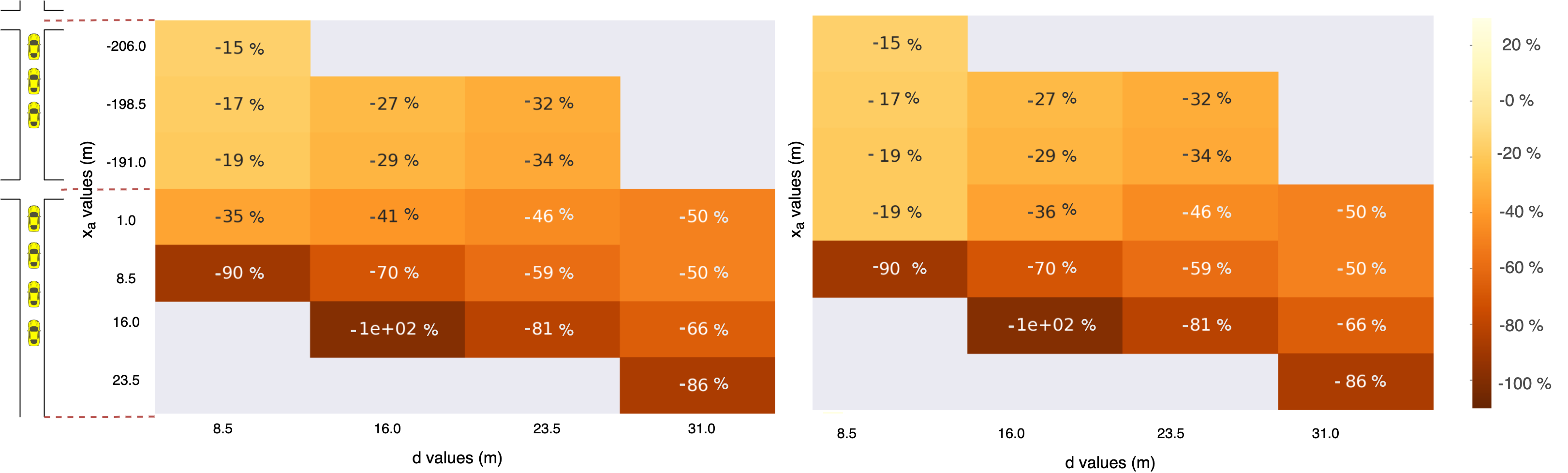}
\caption{Percentage travel time difference of the EMS vehicle under the model-based controller (left) and the DRL controller (right) compared against the oracle. Results are reported with varying $x_a$ and $d$ values. Positive values indicate EMS vehicle travel time under the DRL or model-based controller is smaller than that of oracle while negative values indicate EMS vehicle travel time under the DRL or model-based controller is greater than that of oracle (higher the better).}
\label{fig:ems-opt}
\end{figure*}

\subsection{Optimality Gap of the DRL and Model-based Controllers}
\label{optimal-model}




In this section, we conduct experiments to answer Q3. We define a theoretically optimal controller based on our earlier discussion. We refer to this controller as the 'oracle', in the sense that each vehicle may accelerate or decelerate to the desired speed instantaneously. The main difference between the oracle and the model-based controller is that the oracle produces travel times analytically based on equations provided in Appendix~\ref{jordan-controller} while the model-based controller produces travel times by simulating vehicles in a simulation environment with human-like driving models. We analyze the travel times of the CAV and the EMS vehicle under the DRL controller and the model-based controller against the oracle and summarize the results in Figure~\ref{fig:cav-opt} and Figure~\ref{fig:ems-opt}.

As can be seen in Figure~\ref{fig:cav-opt}, the DRL controller produces CAV travel times that are closer to or better than the oracle. The DRL controller performing better in certain cases indicates the benefit of using a learning-based control method instead of a model-based control method. Both our oracle and model-based controllers are model-based controllers as they assume a model of the vehicle dynamics. Our DRL controller finds better control strategies that can outperform the baselines in real-world like driving environments by leveraging model-free learning that makes no assumptions of the vehicle dynamics or inter-vehicle dynamics. For cases where the oracle outperforms the DRL controller, one can argue that the inter-vehicle dynamics ignored by the oracle provide it with significant flexibility in reducing travel time. However, it is not practically possible to achieve such travel times in real-world like driving environments due to randomness introduced by the human-like driving dynamics. This effect can be seen comparing the Figure~\ref{fig:cav-opt} left and Figure~\ref{fig:cav-opt} right. As can be seen, the model-based controller almost always outperformed by the DRL controller, indicating that the model-based assumptions we make in the model-based controller do not always hold in real-world like simulations. The same phenomenon explains the travel times of the EMS vehicle as illustrated in Figure~\ref{fig:ems-opt}.


\subsection{Computation time}
Regarding the computation time, the model-based controller has a low computation time as it is an analytical solution. DRL, on the other hand, has a high training time ($\sim$8 hours on an Nvidia Volta V100 GPU machine with 5 CPU cores, each having 9GB RAM). However, during inference time DRL controller is as fast as the model-based controller as there is no additional training involved. So, once the DRL controller training is completed, the model-based controller has no computational advantage over the DRL controller.

We would also like to highlight that even though the DRL controller takes more time to train, it has better flexibility in adapting to different scenarios just by simple manipulations of the reward function. The model-based controller, on the other hand, has very limited flexibility in terms of adapting to different scenarios from which it was designed to work and will require careful re-modeling which could be quite time-consuming and will require explicit expert involvement. 

Regarding the factors that affect the training time of DRL agents, we see simulation time as one of the major factors. Our simulation will have to simulate closer to 40 vehicles including EMS vehicles and CAV at any simulation step which could be time-consuming. Second, the hyperparameters used in PPO algorithm can also influence the time to train. For example, hyperparameters like learning rate and entropy coefficient can affect the training time. We manually found the best configuration of hyperparameters that produced the best results with faster training.

\section{Conclusion and future work}

In this work, we consider the corridor clearance problem for emergency vehicles. In particular, we study off-nominal mixed autonomy scenarios in which an EMS vehicle is faced with traffic congestion at signalized intersections. An automated controller designed using deep reinforcement learning (DRL) is proposed to balance the travel time and speed of EMS while minimizing the influence on normal traffic. Our experimental results indicate that learned DRL controller can reduce the travel time of the EMS vehicle as much as previous shockwave-theoretic controllers while even reducing the averse impacts on surrounding traffic.     

Our future work will consider more complex scenarios with grid networks to further evaluate the scalability of the DRL approach. Additionally, rather than treating the control strategy for the traffic signal as fixed, we will consider dynamic traffic signal systems controlled by DRL for the grid networks. The traffic signal controller may coordinate with the DRL module designed for vehicles to further assist the EMS vehicles. The simulation results presented in this paper can be used as a benchmark for evaluating collaborative designs among heterogeneous autonomous agents (e.g., traffic signals and CAVs).

\section*{Acknowledgment}

The authors would like to thank the MIT Supercloud for providing support on the experiment platform used in the simulation. The authors extend their gratitude to the anonymous reviewers whose suggestions significantly enhanced the quality of the paper.

\bibliographystyle{IEEEtran}
\bibliography{root.bib}

\newpage

\section{Biography Section}

\vspace{11pt}
\vskip -2\baselineskip plus -1fil
\begin{IEEEbiography}[{\includegraphics[width=1in,height=1.25in,clip,keepaspectratio]{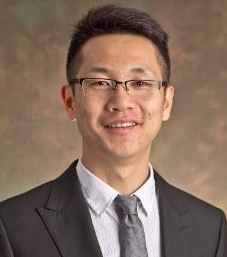}}]{Dajiang Suo} is an Assistant Professor at Arizona State University. Before that, he was a research scientist at the Massachusetts Institute of Technology. Suo obtained a Ph.D. in Mechanical Engineering from MIT in 2020. Suo holds a B.S. degree in Mechatronics Engineering, and an S.M. degree in Computer Science and Engineering Systems. His research interests include the Internet of Things, connected vehicles, cybersecurity, and RFID.

Before returning to school to pursue PhD degree, Suo was with the vehicle control and autonomous driving team at Ford Motor Company (Dearborn, MI), working on the safety and cyber-security of automated vehicles. He also serves on the Standing Committee on Enterprise, Systems, and Cyber Resilience (AMR40) at the Transportation Research Board.
\end{IEEEbiography}

\vspace{11pt}

\begin{IEEEbiography}[{\includegraphics[width=1in,height=1.25in,clip,keepaspectratio]{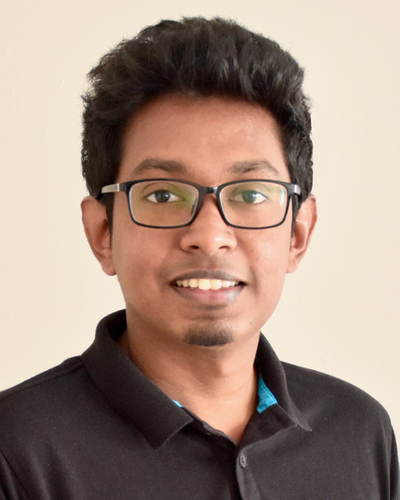}}]{Vindula Jayawardana} received the B.S. degree in computer science and engineering from the University of Moratuwa, Sri Lanka, in 2018 and the M.S degree in computer science from MIT in 2022. He is currently working toward his Ph.D. degree in computer science at the Laboratory for Information \& Decision Systems at MIT.

His broader research interests focus on advancing the understanding of learning for control. His current work focuses on making multi-agent reinforcement learning seamlessly generalize across problem variations with applications in planning for autonomous vehicles.

\end{IEEEbiography}

\vspace{11pt}

\begin{IEEEbiography}[{\includegraphics[width=1in,height=1.25in,clip,keepaspectratio]{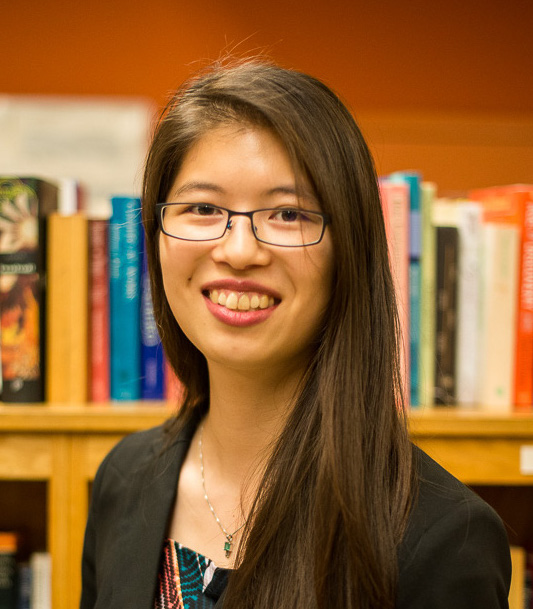}}]{Cathy Wu} received the B.S. and M.Eng. degrees from the Massachusetts Institute of Technology (MIT), Cambridge, MA, USA, in 2012 and 2013, respectively, and the Ph.D. degree from the University of California Berkeley,  , all in electrical engineering and computer sciences.

She was a Postdoc with the Microsoft Research AI. She is an Assistant Professor with MIT in LIDS, CEE, and IDSS. She studies the technical challenges surrounding the integration of autonomy into societal systems. Her research interests include machine learning and mobility. Prof. Wu was a recipient of several awards, including the 2019 IEEE Intelligent Transportation Systems Society (ITSC) Best Ph.D. Dissertation Award, 2018 Milton Pikarsky Memorial Dissertation Award, and the 2016 IEEE ITSC Best Paper Award, and has appeared in the press, including Wired and Science.
\end{IEEEbiography}

\newpage

\section{Appendix}
This section discusses the necessary modifications needed to extend the single-intersection model-based controller to a two-intersection model-based controller. Unlike the DRL approach (presented in Section V-B), which mainly involves shaping the reward coefficient in the proposed Markov Decision Process, the model-based approach proves to be time-consuming and unscalable. For the latter, any changes in the inter-vehicle dynamics, queue dissipation rate, or traffic signal timing necessitate model reformulations when the EMS vehicle needs to travel through more than one intersection.

\subsection{Shock-wave theoretic approach for model-based control}
\label{jordan-controller}
We present a simplified realization of the optimal control formulation in Section IV-A. The realization is based on shock-wave theory and was originally discussed in~\cite{jordan2013path}. Due to its inherent limitations discussed in Section V-A, the shock-wave theory-based approach can only target a special case of the optimal control problem. Our goal is to use it to highlight the increased complexity in model reformulations for the model-based method as we increase the number of intersections the EMS vehicle needs to move through. 

\begin{figure*}[tb!]
\centering
\begin{subfigure}{0.44\textwidth}
  \centering
  \includegraphics[width=\textwidth]{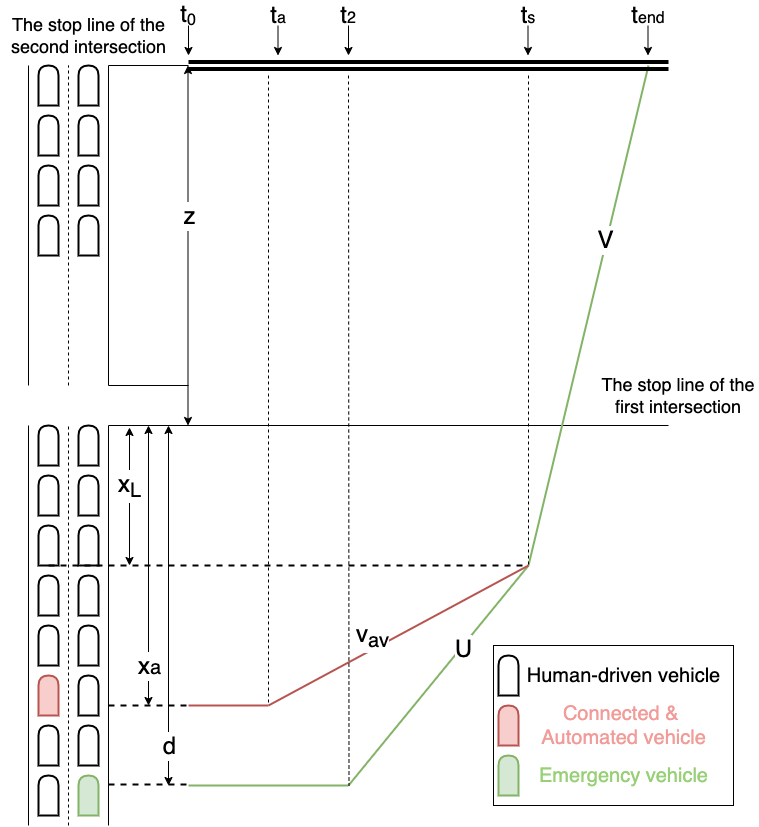}
  \caption{The CAV is located at the position $x_a$ upstream of the optimal splitting point $x_L$.}
  \label{fig:sameedge}
\end{subfigure}
~
\begin{subfigure}{0.45\textwidth}
  \centering
  \includegraphics[width=\textwidth]{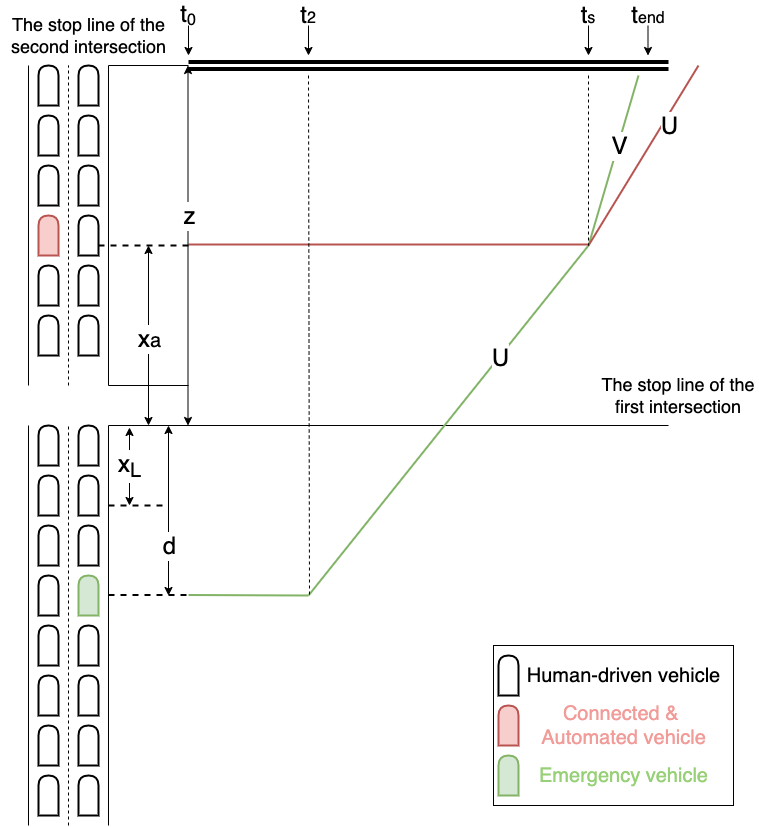}
  \caption{The CAV is located at the position $x_a$ downstream of the optimal splitting point $x_L$.}
  \label{fig:diffEdge}
\end{subfigure}
\caption{CAV-assisted corridor-clearance scenarios where the EMS needs to travel through two intersections to get to its destination.}
\label{fig:two_Intersection}
\end{figure*}

A key variable used for designing the model-based controller is the optimal splitting point $x_L$, as shown in Fig.~\ref{fig:singleInter} and \ref{fig:two_Intersection}. It is optimal in the sense that the splitting point minimizes the EMS vehicle's travel time to the intersection. Additionally, the EMS vehicle, after switching to the left lane, can travel at its desired emergency speed without being disrupted by the leading vehicle (orange color in Fig.~\ref{singleInter_highpenetration}) until reaching the intersection, as shown in the time-space diagram in Fig.~\ref{singleInter_highpenetration}.

At time $t_0$, the traffic signal (north-south direction) turns green, and therefore, the queues start to discharge, and the CAV and EMS vehicle are located at position $x_a$ and $d$, respectively. It is worth mentioning that $t_0$ has no influence on the starting position of the EMS vehicle $d$, which is already fixed at the end of the previous traffic signal cycle. The derivation of the optimal splitting point $x_L$ is based on the stationary assumption that a constant shockwave speed $w$, which is the speed of queue discharging, can be derived using historical data from road detectors~\cite{wang2013design,liu2009real}.
The calculation uses $w$ to compute the waiting time $t_1$ of the vehicle in front of a candidate splitting point (represented by the orange bar) and the waiting time $t_2$ of the EMS vehicle, derived in Eq.~\ref{naiveJordan}a and \ref{naiveJordan}b.

Since the time until the EMS vehicle arrives at the splitting point $t_s$ can be calculated by using Eq.~3c, we can derive the travel time of the EMS vehicle until it gets to the intersection $t_{ev}$ as seen in Eq.~3d. The EMS vehicle will move at the same speed as other vehicles in the queue (i.e., $U$) and then travel at its desired speed $V$. Also, it should be noted that the speed of traffic moves faster than the queue dissipation rate. Therefore, we can assume that $w \leq U \leq V$. On the other hand, the travel time for the vehicle preceding the optimal point to reach the intersection is given in Eq.~3e. The optimal splitting point $x_L$ can then be derived from the simultaneous solution of Eq.~\ref{naiveJordan}d and \ref{naiveJordan}e, which is given in Eq.~\ref{naiveJordan}f.

\begin{subequations}\label{naiveJordan}
\begin{align}
t_1 = \quad & \frac{x_L}{w} \\
t_2 = \quad & \frac{d}{w} \\
t_s = \quad & t_2+\frac{d-x_L}{U}   \\
t_{ev} = \quad & t_s + \frac{x_L}{V} \\
t_{pre} = \quad & t_1 + x_L/U \\
x_L = \quad & d\frac{w^{-1}+U^{-1}}{w^{-1}+2U^{-1}-V^{-1}}
\end{align}
\end{subequations}

\subsubsection{The Model-based controller for single-intersection scenarios}
Consider that only one CAV is present in the adjacent lane of the EMS. We then need to determine the optimal strategy for the longitudinal control of the CAV based on whether or not the actual location of the CAV (e.g., $x_a$) is less than the (theoretical) optimal splitting point ($x_L$), as shown in Fig.~\ref{fig:singleInter}. It should be noted that the model-based controller derived here must be sub-optimal as the EMS vehicle does not have the flexibility of choosing $x_L$ as the optimal splitting point due to low CAV penetration.

Therefore, for the case where $x_a \leq x_L$ (i.e., when the CAV starts from a location closer to the intersection than the optimal splitting point), the best strategy for the CAV is to keep stationary until the EMS vehicle has switched to the left lane (i.e., CAV applying the brake such that $v_{cav} = 0$). 

The travel time of the CAV in this scenario is given in eq.~\ref{eq:single_T_cav}. The first term represents the time it takes for the queue on the right lane to discharge while the second terms denotes the time for the EMS to move to the position $x_a$ where the CAV locates. After the EMS finishes lane-changing, it can speed up and travel at its desired speed $V$ until it crosses the intersection, as represented by the third term.

\begin{equation}\label{eq:single_T_cav}
T_{cav} = \frac{d}{w}+\frac{d-x_a}{w}+ \frac{x_a}{U}
\end{equation}

For the case where $x_a \geq x_L$ (i.e., when the CAV is at a location further away from the intersection than the optimal splitting point), Eq.~\ref{eq:single_T_cav} shows the travel time. The first term represents the time it takes for the queue to discharge, while the second term denotes the time for the vehicle preceding the CAV to cross the intersection.

\begin{equation}\label{eq:single_T_cav}
T_{cav} =   \frac{x_a}{w}+ \frac{x_a}{U}
\end{equation}

As mentioned earlier, the derivation given above only optimizes the travel time of the EMS without considering the CAV. Empirically, the CAV can still move at a speed $v_{cav}$ such that the vehicles behind the CAV can move through the second intersection faster without being blocked by the traffic signal. In our simulation experiments, we change the value of $v_{cav}$ with fine-tuning but make sure that the CAV will not reach the optimal splitting point $x_L$ earlier than the EMS vehicle, as seen in Eq.~\ref{jordancondition_1}. 

\begin{equation}\label{jordancondition_1}
\frac{x_a - x_L}{v_{cav}} > t_s - t_{a}
\end{equation}
where $t_a$ denotes the time when the CAV is first able to move, based on the queue discharge.

To derive $t_a$, we can apply the shockwave speed to the time-space diagram in Fig.~\ref{fig:singleInter}. Specifically, $t_a$ is equal to the distance between the intersection and the CAV $x_a$ divided by the queue discharging speed $w$, as seen in Eq.~\ref{jordancondition_2}.

\begin{equation}\label{jordancondition_2}
t_a = \frac{x_a}{w}
\end{equation}


\subsubsection{Model reformulations for two-intersection scenarios}


On the surface of it, the derivation of the model-based controller for two-intersection looks similar to single-intersection scenarios. However, the status of the queue near the second intersection can influence the movement of the EMS after it switches to the left lane and also the strategy for the model-based controller, as shown in Fig.~\ref{fig:two_Intersection}.

First, the movement of the EMS vehicle and the preceding vehicles that start upstream of the first intersection can be hindered by the queue downstream. This occurs when the length of the queue downstream of the first intersection is above a certain threshold, or the individual discharge rate of the queue downstream of the first intersection is smaller than the queue upstream such that the queue can not be discharged in time before the EMS reaches the tail of the queue.

Second, for the case in which the CAV is downstream of the optimal splitting point discussed earlier, as shown in Fig.~\ref{fig:diffEdge}, there is an exception when the CAV may choose to go across the second intersection without waiting for the EMS vehicle. Specifically, if the queue downstream of the first intersection can be dissipated in time, the CAV should start to move when the traffic signal for the second intersection turns green, as doing this will not hinder the movement of the EMS vehicle. The prerequisite condition is given in Eq.~\ref{eq:jordan_diffedge_special}. However, without real-time data on the rate at which the queue is clearing and the timing of the traffic signals, the model-based controller faces challenges in determining when the conditions in Eq.~\ref{eq:jordan_diffedge_special} are met. This necessary information remains inaccessible unless there's a vehicle-to-infrastructure communication system in place.

\begin{equation}\label{eq:jordan_diffedge_special}
\frac{z-{x_a}}{w}+\frac{z-{x_a}}{U} \leq  \frac{d}{w}+\frac{d}{V}
\end{equation}




\vfill

\end{document}